\documentclass{article} 
\pdfoutput=1
\usepackage[margin=1in]{geometry}
\RequirePackage{natbib}  

\usepackage{amsmath,amsfonts,bm}

\def\eqref#1{equation~\ref{#1}}

\def\1{\bm{1}}

\DeclareMathAlphabet{\mathsfit}{\encodingdefault}{\sfdefault}{m}{sl}
\SetMathAlphabet{\mathsfit}{bold}{\encodingdefault}{\sfdefault}{bx}{n}

\def\gO{{\mathcal{O}}}

\usepackage{multirow}
\usepackage{caption}
\usepackage{comment}

\usepackage{url}

\UseRawInputEncoding
\usepackage{xcolor}
\definecolor{tabblue}{HTML}{5555CC}
\usepackage[hidelinks,colorlinks=true,linkcolor=tabblue,citecolor=tabblue]{hyperref}
\usepackage{graphicx}  
\usepackage{booktabs}  
\usepackage{amssymb}
\usepackage{pifont}

\usepackage[normalem]{ulem}
\useunder{\uline}{\ul}{}

\usepackage[utf8]{inputenc}  

\UseRawInputEncoding  

\title{The Hedgehog \& the Porcupine: Expressive Linear Attentions with Softmax Mimicry}

\author{Michael Zhang, Kush Bhatia, Hermann Kumbong and Christopher R\'{e} \\
\vspace{-0.35cm}
\\
{\ Department of Computer Science, Stanford University }\\
\vspace{-0.35cm}
\\
\texttt{\{mzhang,kushb,chrismre\}@cs.stanford.edu}, \texttt{kumboh@stanford.edu},
}

\date{}

\newcommand{\name}{Hedgehog}

\newcommand{\cmark}{\ding{51}}%
\newcommand{\xmark}{\ding{55}}%

\newcommand{\eg}{\textit{e.g.,}}

\newcommand{\update}[1]{{{{#1}}}}

\usepackage{enumitem}  
\usepackage{wrapfig,lipsum}

\usepackage{listings}  
\definecolor{codegreen}{rgb}{0,0.6,0}
\definecolor{codegray}{rgb}{0.5,0.5,0.5}
\definecolor{codepurple}{rgb}{0.58,0,0.82}
\definecolor{backcolour}{rgb}{0.95,0.95,0.95}

\lstdefinestyle{mystyle}{
    backgroundcolor=\color{backcolour},   
    commentstyle=\color{codegreen},
    keywordstyle=\color{magenta},
    numberstyle=\tiny\color{codegray},
    stringstyle=\color{codepurple},
    basicstyle=\ttfamily\footnotesize,
    breakatwhitespace=false,         
    breaklines=true,                 
    captionpos=b,                    
    keepspaces=true,                 
    numbers=left,                    
    numbersep=5pt,                  
    showspaces=false,                
    showstringspaces=false,
    showtabs=false,                  
    tabsize=2,
    escapeinside={<@}{@>}
}

\begin{document}

\maketitle

\begin{abstract}
    Linear attentions have shown potential for improving Transformer efficiency, reducing attention's quadratic complexity to linear in sequence length. This holds exciting promise for (1) training linear Transformers from scratch, (2) ``finetuned-conversion'' of task-specific Transformers into linear versions that recover task performance, and (3) ``pretrained-conversion'' of Transformers such as large language models into linear versions finetunable on downstream tasks. However, linear attentions often underperform standard softmax attention in quality. To close this performance gap, we find prior linear attentions lack key properties of softmax attention tied to good performance: low-entropy (or ``spiky'') weights and dot-product monotonicity. We further observe surprisingly simple feature maps that retain these properties and match softmax performance, but are inefficient to compute in linear attention. We thus propose \name{}, a learnable linear attention that retains the spiky and monotonic properties of softmax attention while maintaining linear complexity. \name{} uses simple trainable MLPs to produce attention weights mimicking softmax attention. Experiments show \name{} recovers over 99\% of standard Transformer quality in train-from-scratch and finetuned-conversion settings, outperforming prior linear attentions up to 6 perplexity points on WikiText-103 with causal GPTs, and up to 8.7 GLUE score points on finetuned bidirectional BERTs. \name{} also enables pretrained-conversion. Converting a pretrained GPT-2 into a linear attention variant achieves state-of-the-art 16.7 perplexity on WikiText-103 for 125M subquadratic decoder models. We finally turn a pretrained Llama-2 7B into a viable linear attention Llama. With low-rank adaptation, \name{}-Llama2 7B achieves 28.1 higher ROUGE-1 points over the base standard attention model, where prior linear attentions lead to 16.5 point drops.
\end{abstract}
\section{Introduction}
\label{sec:intro}

Linear attentions are promising methods for improving Transformer efficiency. By replacing the softmax of attention's query and key dot products with kernel function feature maps, linear attentions reduce attention's time and space complexity from $\mathcal{O}(n^2d)$ to $\mathcal{O}(ndd')$ where $n$ is sequence length, $d$ is head dimension and $d'$ the feature map dimension~\citep{katharopoulos2020transformers, choromanski2020rethinking, peng2021random, xiong2021nystromformer, schlag2021linear}. For typical Transformer settings, \eg{} with head dimension $=$ 64 and sequence lengths at 512 to 32K, this quadratic-to-linear scaling can result in significant speed and memory improvements (Fig.~\ref{fig:scaling}). 
As drop-in alternatives to popular softmax attention~\citep{vaswani2017attention}, linear attentions not only improve Transformer efficiency when training new models from scratch but can also improve inference efficiency by converting pretrained Transformers into corresponding linear variants~\citep{kasai-etal-2021-finetuning, mao-2022-fine}. 
Linear attention enables efficient Transformers in a variety of regimes:
\begin{itemize}[leftmargin=*]
    \item \textbf{Training-from-scratch}: training Transformer models with linear attention with the goal of matching standard Transformer performance, \eg{} as tested on benchmarks such as Long Range Arena (LRA) classification~\citep{tay2021long} and WikiText-103 language modeling ~\citep{merity2017pointer}.
    \item \textbf{Finetuned-conversion}: swapping the attentions of task-specific Transformers and finetuning them to convert existing models into linear versions, with the goal to recover original task performance with improved efficiency~\citep{kasai-etal-2021-finetuning, mao-2022-fine}.
    \item \textbf{Pretrained-conversion}: doing the same as finetuned-conversion but for pretrained Transformers such as large language models (LLMs), \eg{} to transfer to new tasks and longer contexts.
\end{itemize}

Unfortunately, existing linear attention mechanisms typically fail to match softmax attention in modeling quality. When training from scratch, linear attentions achieve 4-6 worse perplexity (ppl) than softmax attention on standard benchmarks such as WikiText-103~\citep{schlag2021linear, irie2021going, fu2023hungry}, the equivalent gap between 125M and 255M Transformers~\citep{dai-etal-2019-transformer}.
When converting finetuned models, linear attention models require additional quadratic attention modules to close the gap~\citep{kasai-etal-2021-finetuning, mao-2022-fine}.
One might worry that such gaps are fundamental; for example, recent theory using the Strong Exponential Time Hypothesis (SETH) showed that high-quality truly subquadratic algorithms to approximate softmax attention may be impossible with large sequence length $n$~\citep{alman2023fast, keles2023computational}.

We begin by empirically studying why this performance gap exists between standard softmax and proposed linear attentions. 
We identify two simple properties for softmax attention which prior linear attentions lack: 1) low-entropy ``spikyness'' and 2) dot-product monotonicity. We hypothesize that the quality gap in linear attentions corresponds with lacking these two properties: \vspace{-2mm}
\begin{itemize}[leftmargin=*]
    \item \textbf{Low-entropy ``spikyness''}: Intuitively, we want attentions that attend to relevant tokens while ignoring irrelevant ones via their query-key interactions. 
    We observe these low-entropy or ``spiky'' attention-weight distributions in standard Transformer attention but not prior linear attention maps---where spikes enabled via the scaled dot-product softmax are lost via other feature maps (Fig.~\ref{fig:entropy_viz})---and find this strongly corresponds to Transformer performance (Fig.~\ref{fig:results_ar}).
    \item \textbf{Dot-product monotonicity}: This property requires that attention weights increase as the dot products of their corresponding queries and keys increase. Intuitively, the lack of this monotonicity can produce unstable gradients during training and finetuning, where increasing the query-key dot product can result in decreasing the attention weight the other way (and vice versa).
\end{itemize}
As a first step to recover these properties, we explore simple feature maps---such as low-degree Taylor polynomial approximations to the $\exp()$ function---that satisfy the above two properties (albeit in restricted regimes of bounded query-key dot products). In practice, we find that queries and keys are often bounded, resulting in linear attentions that recover softmax attention's spikiness, monotonicity, and subsequent performance.
Unfortunately, while technically linear in sequence length, these polynomial feature maps remain inefficient to compute. They take $\gO(nd^{p+1})$ time and space, and we find degree $p \geq 2$ necessary for performance. 

\begin{figure}[t] 
    \centering
    \includegraphics[width=1\textwidth]{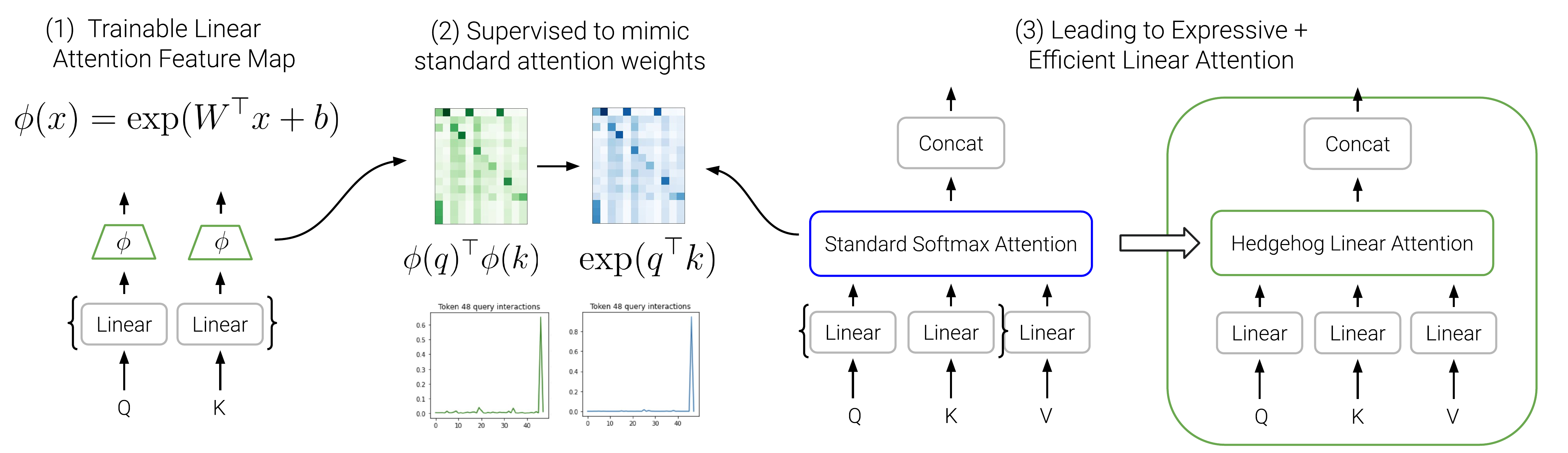}
    \caption{\name{} learns a trainable linear attention feature map designed to mimic standard attention, resulting in expressive yet efficient linear attentions for various Transformer training settings}
    \label{fig:pull_fig}
    \vspace{-0.5cm}
\end{figure}

We thus propose \name{}, an efficient-to-compute \emph{learnable} linear attention trained to capture the spiky and monotonic softmax properties. Unlike prior works that propose a specific kernel function~\citep{katharopoulos2020transformers, choromanski2020rethinking, qin2022cosformer} and our polynomial feature maps, we learn these feature maps as single-layer MLPs specifically \emph{trained to match} softmax attention weights. By mapping from $\mathbb{R}^d \mapsto \mathbb{R}^d$, we maintain prior linear attentions' $\gO(nd^2)$ complexity. However, training these mappings via softmax attention weights as cross-entropy soft-labels, we find \name{} can match softmax attention weights with much higher fidelity (Fig.~\ref{fig:attention_matching}), producing low-entropy and monotonic weights that match standard attention performance quality.

We validate experimentally that \name{}'s improved expressivity translates to closing the softmax attention performance gap in the three regimes mentioned above:
\begin{itemize}[leftmargin=*]
    \item \textbf{Training-from-scratch}: we find \name{} matches Transformers on standard attention benchmarks such as Long Range Arena (LRA)~\citep{tay2021long} task, and closes the linear attention gap by 68.6\% on WikiText-103 language modeling (improving up to 6 ppl).
    \item \textbf{Finetuned-conversion}: we find \name{} recovers $>$99\% of original model performance on average across bidirectional encoder-only 110M BERT-base models finetuned on GLUE and causal decoder-only 125M GPT models finetuned on Wikitext-103.
    %
    \item \textbf{Pretrained-conversion}: we find \name{} enables effective transfer to new tasks and efficient scaling to longer contexts, while frequently outperforming modern subquadratic sequence architectures by linearizing existing pretrained Transformers. A 125M \name{}-GPT-2 finetuned on Wikitext-103 achieves a new state-of-the-art 16.7 ppl for subquadratic models of the same size. 
\end{itemize}

Finally, we demonstrate that \name{} can be scaled up to modern large language models; we convert pretrained Llama-2 7B into a viable linear attention Llama. 
With low-rank adaptation, \name{}-Llama2 7B achieves up to 28.1 higher ROUGE-1 points over the base standard attention model. In contrast, prior linear attentions result in models that struggle to produce coherent text (with 16.5 ROUGE-1 point drops).
\section{Preliminaries and Related Work}
\label{sec:preliminaries}

We provide background on attention computation, describe kernel feature based linear attentions, and finally provide details on existing linear attention mechanisms proposed in the literature.

\textbf{Attention setup}. Let $\{\bm{q}_i\}_{i=1}^n$, $\{\bm{k}_i\}_{i=1}^n$, $\{\bm{v}_i\}_{i=1}^n$ denote the set of queries, keys, and values, with individual elements in $\mathbb{R}^d$. Let $n$ denote sequence length and $d$ denote head dimension. We compute attention outputs $\bm{y}_i \in \mathbb{R}^d$ by first computing similarities between each $\bm{q}_i$ and every $\bm{k}_j$; for causal attention we compute these similarities for $j \leq i$. The vanilla Transformer attention computes these similarities using the softmax dot products~\citep{vaswani2017attention}:
{\small
\begin{equation}
\label{eq:standard_attn}
    \bm{y}_i = \sum_{j = 1}^{i}\text{sim}(\bm{q}_i, \bm{k}_j) \bm{v}_j,
    \quad \text{where}\quad
    \text{sim}(\bm{q}_i, \bm{k}_j) = \frac{\exp(\bm{q}_i^\top \bm{k}_j /\sqrt{d}) }{\sum_{m = 1}^{i} \exp(\bm{q}_i^\top \bm{k}_m /\sqrt{d})}\;.
\end{equation}
}
While very expressive, computing attention via Eq.~\ref{eq:standard_attn} 
for all $\{\bm{y}_i\}_{i=1}^n$ requires $\mathcal{O}(n^2 d)$ time and memory, making this inefficient for long sequences. 
To improve efficiency without sacrificing quality, we thus want alternative \emph{linear attention} maps which maintain standard attention's expressivity. 

\textbf{Linear attention and kernel functions}. 
Observe that the $\exp(\cdot)$ in Eq.~\ref{eq:standard_attn} can be viewed as a kernel function, which ~\cite{tsai-etal-2019-transformer, katharopoulos2020transformers} show can be replaced 
in general with 
    $\mathcal{K}(\bm{x}, \bm{x}') = \phi(\bm{x})^\top \phi(\bm{x}')$. 
Here $\phi: \mathbb{R}^d \mapsto \mathbb{R}^{d'}$ is a feature map applied to each vector. 
We can thus compute attention in \emph{linear} time and space over the sequence length $n$, seen by rewriting Eq.~\ref{eq:standard_attn} as:
{\small
\begin{equation}\label{eq:linear_attention}
    \bm{y}_i = \frac{\phi(\bm{q}_i) \sum_{j = 1}^i \big( \phi(\bm{k}_j)^\top \bm{v}_j \big)
    }{\phi(\bm{q}_i) \sum_{j = 1}^i \phi(\bm{k}_j)}\;.
\end{equation}
}

\textbf{Prior feature maps}. From the previous section, we observe that  
linear attentions are promising directions for improving Transformer efficiency at both training and inference time. Numerous prior works have proposed feature maps $\phi$ aiming to remain more efficient (where linear attention is desirable to standard attention if $d' < n$), while still being expressive and stable to train. These range from $\phi$ ensuring positive attention weights, \eg{} via $1 + \text{ELU}$~\citep{katharopoulos2020transformers} or $\text{ReLU}$~\citep{kasai-etal-2021-finetuning}, to softmax or Gaussian kernel approximations via randomized features~\citep{rahimi2007rff,choromanski2020rethinking,peng2021random,choromanski2021hybrid,zheng2023efficient} or low-rank approximations~\citep{xiong2021nystromformer, chen2021skyformer}.
\section{Improving Linear Attention via Spiky and Monotonic Weights} 
\label{sec:analysis}
We begin by identifying two key properties of attention weights which we hypothesize are essential for good performance quality. The first, \emph{low-entropy spikyness}, requires that the attention map is able to capture effectively capture sparse relevant tokens in a sequence. The second, \emph{monotonicity over query-key dot products}, requires the attention map to increase with increasing dot products, and allows for smooth conversion of pretrained Transformers into linear variants. 

\begin{figure}[t] 
    \centering
    \includegraphics[width=1\textwidth]{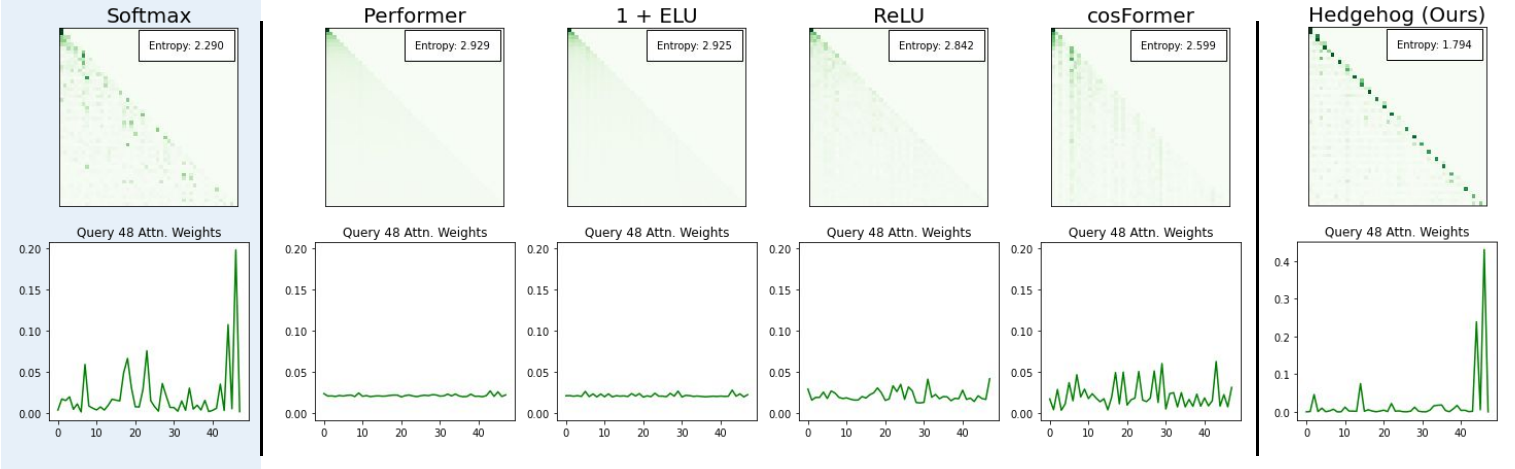}
    \caption{\textbf{Attention weight spikiness}. \update{(Plots 1 - 5):} Softmax attention results in lower entropy and ``spiky'' selective weighting compared to prior linear attentions (training from scratch on associative recall (Sec.~\ref{sec:analysis_experiments})). \update{(Plot 6): By training to mimic softmax attention, our proposed \name{} recovers this spikiness as a linear attention, corresponding with improved  performance (Sec.~\ref{sec:results}).} 
    }
    \label{fig:entropy_viz}
\end{figure}

\subsection{Properties for Expressive Attention Maps}
\label{sec:analysis_def}
\begin{figure}[t] 
    \centering
    \includegraphics[width=1\textwidth]{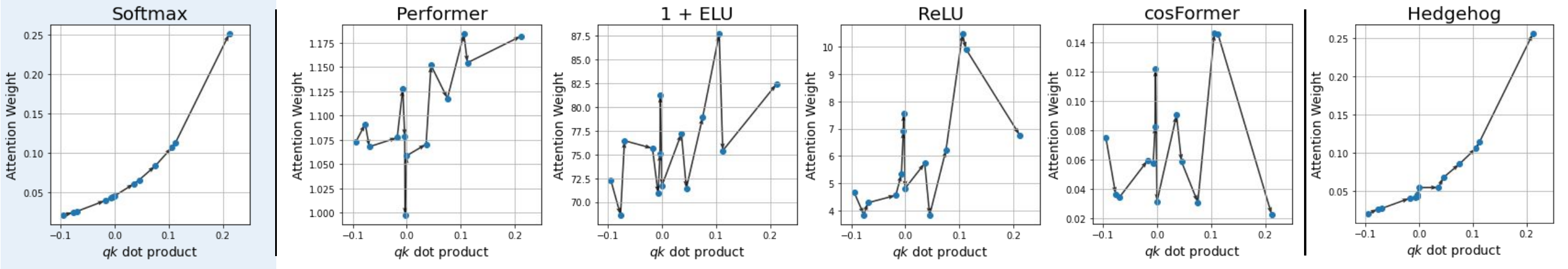}
    \caption{\textbf{Attention weight monotonicity}. \update{(Plots 1 - 5):} In contrast to softmax attention, prior linear attentions are not smoothly monotonic over trained query-key dot products, resulting in poor performance when converting BERT models by replacing attentions (Table ~\ref{table:sec3_train_scratch_result}). \update{(Plot 6): \name{} recovers this monotonicity, and thus recovers 99\% of BERT performance after conversion  (Table~\ref{table:ft_conversion}).}}
    \label{fig:monotonic_viz}
\end{figure}

Here we describe the spiky and monotonic properties hypothesized for desirable linear attention.
We note these add to past observations for more performant linear attentions, including positive attention weights~\citep{katharopoulos2020transformers}, orthogonal features~\citep{choromanski2020rethinking, irie2021going}, or locality (upweighting nearby values)~\citep{qin-etal-2022-devil, qin2022cosformer}. \update{We validate these properties among past linear attentions in Sec.~\ref{sec:analysis_experiments}, and preview how our proposed \name{} linear attention recovers these properties in correspondence with improved performance (Sec.~\ref{sec:results}) in Fig.~\ref{fig:entropy_viz},~\ref{fig:monotonic_viz}.}

\begin{table}[t]
\vspace{-0.5cm}
\centering
\resizebox{\linewidth}{!}{
\begin{tabular}{@{}lccccccc@{}}
\toprule
                                            & BERT-FT & 1 + ELU & ReLU & Performer & cosFormer & $\exp$(t = 1) & $\exp$(t = 2) \\ \midrule
Matthew's correlation & \textbf{58.8}    & {28.1}    & {39.5} & {24.7}      & {39.9}      & {45.9}       & {50.0}         \\
\bottomrule
\end{tabular}
}
\caption{Finetuned-conversion performance of BERT finetuned on CoLA (BERT-FT), using prior linear attentions. \update{With poor monotonicity (Fig.~\ref{fig:monotonic_viz}), prior methods fail to recover performance.}} 
\label{table:sec3_train_scratch_result}
\vspace{-0.5cm}
\end{table}

\paragraph{Low-entropy spikiness.} Intuitively, one source of attention's effectiveness is its ability to selectively upweight relevant tokens in a sequence. This is a popular interpretation visualized in various Transformer architectures and settings ranging from encoder-decoder language translation~\citep{bahdanau2014neural} to ViT image segmentation~\citep{dosovitskiy2020image, caron2021emerging}. 
Mechanically, the softmax over query-key dot products exponentiates relative similarities between a query and each key, quantified via low-entropy or ``spiky'' attention weight distributions (Fig.~\ref{fig:entropy_viz}).
 
Linear attention maps work by replacing the softmax with the normalized dot product of alternate feature maps (Eq.~\ref{eq:linear_attention}). With existing feature maps, we find the resulting attention weights can result in much higher entropy or more uniform distributions. This is true even for methods designed to approximate the softmax under mean-squared error bounds~\citep{choromanski2020rethinking} (Performer, Fig.~\ref{fig:entropy_viz}) or imposed locality~\citep{qin2022cosformer} (cosFormer, Fig.~\ref{fig:entropy_viz}). This uniformity in attention weights reduces the modeling capacity of linear attentions leading to worse performance quality.

\paragraph{Monotonicity over query-key dot products.} This property requires that the attention maps are monotonic over query-key dot products: when the dot product increases (decreases), the attention weight increases (decreases). In Fig.~\ref{fig:monotonic_viz}, we observe that while softmax attention exhibits this monotonicty (first subplot), the existing linear attentions do not. We believe this can cause training issues after swapping attentions due to conflicting gradients between attentions and original model parameters.
In Fig.~\ref{fig:monotonic_viz}, trying to upweight attentions by increasing product similarity can actually result in 
\emph{decreased} attention weights. Later in Sec~\ref{sec:analysis_experiments}, we find this corresponds to failing to recover original performance when converting finetuned Transformers.

\subsection{Explaining the Linear Attention Performance Gap}
\label{sec:analysis_experiments}
We validate the two properties introduced above by showing that (1) lacking spikiness corresponds to significantly worse performance when training from scratch, and (2) lacking spikiness and monotonicity corresponds to failing to recover performance when converting finetuned models.

\textbf{Training from scratch}. We compare various Transformers' abilities to solve Associative Recall (AR)~\citep{ba2016using}, a next-token prediction task previously studied as a proxy for language modeling capability~\citep{olsson2022context}. AR tests how well a model can recall specific content in an input sequence, structured as a list of key-value pairs which ends in a key (Table~\ref{table:associative_recall}). 

\begin{wrapfigure}{r}{0.4\textwidth}
\vspace{-0.75cm}
  \centering
 \includegraphics[width=1\linewidth]{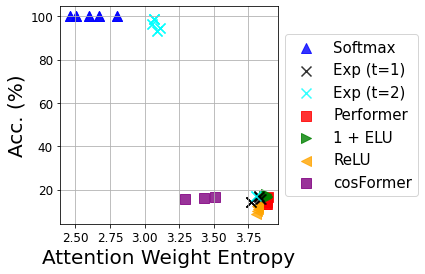}
\vspace{-0.5cm}
\captionof{figure}{Associative recall performance strongly corresponds to lower attention entropy; present in softmax attention but not prior linear variants.}
\label{fig:results_ar}
\vspace{-0.35cm}
\end{wrapfigure}
As a control for evaluating our hypothesis, we also consider a simple feature map designed to induce ``spikiness'' but not monotonicity: $\phi_t(x) = \exp(x \cdot t)$, which applies a temperature-$t$ scaled exponential element-wise.

In Fig.~\ref{fig:results_ar}, we observe a strong correspondence between low-entropy attention weights and AR accuracy. While softmax attention solves the AR task perfectly, prior linear attentions struggle to achieve even 20\% accuracy, at the same time obtaining much larger attention weight entropies. As further support to our hypothesis, we see that while the exponential map $\phi_1$ fails AR and produces similarly high entropy attention weights, increasing spikiness with $t = 2$ actually solves the task.

\textbf{Finetuned-conversion}. We next compare how various linear attentions perform at recovering original softmax attention performance for finetuned-conversion. We adopt the procedure in~\cite{kasai-etal-2021-finetuning}, which takes a Transformer already finetuned on a specific task, swaps the attention layers with a linear attention variant, and further finetunes the entire model on the same task.

For this setting, we evaluate with a BERT-base-uncased model~\citep{devlin2018bert} finetuned on the Corpus of Linguistic Acceptability (CoLA) task~\citep{warstadt2019neural}, where the goal is to classify whether a sentence is grammatically correct. We compare the performance of the original (softmax attention) BERT model\footnote{\href{https://huggingface.co/JeremiahZ/bert-base-uncased-cola}{https://huggingface.co/JeremiahZ/bert-base-uncased-cola}} with the linear attention converted models. In Table~\ref{table:sec3_train_scratch_result}, we find that just as no linear attention smoothly captures monotonicity over the trained model's query-key dot products, no linear attentions fully recover the original finetuned BERT's Matthew's correlation of 58.8. This includes the spiky $\phi_2$ feature map which was sufficient in the training-from-scratch regime.
\section{\name: Expressive Linear Attention via Softmax Mimicry}
\label{sec:method}
We present \name{}, a simple, efficient, and expressive feature map trained to mimic softmax attention. \name{} is predicated by (1) there existing linear attention approximations to the softmax that recover the spiky and monotonic properties of standard attention in practice, and (2) that we can efficiently compute similar approximations efficiently.

In Sec.~\ref{sec:softmax_approx}, we \update{motivate \name{} and show that (1) is possible by revisiting low-degree Taylor polynomials. We find that for linear attention, the Taylor exponential works as a surprisingly simple feature map, recovering spikiness and monotonicity while matching standard Transformer performance. 
Unfortunately, we also find it introduces its own issues, where the feature map results in large query and key dimensions and becomes inefficient to compute.}
In Sec. \ref{sec:hedgehog_method}, to thus overcome these challenges,  we propose and describe \name{}, a \emph{trainable} linear attention trained to mimic softmax attention. In Sec. \ref{sec:hedgehog_benchmarking}, we show how this enables similar spiky and monotonic properties to the softmax and Taylor exponential attentions, while retaining past linear attentions' efficiency.

\subsection{\update{Simple Polynomial Approximations to Softmax Attention}}
\label{sec:softmax_approx}
From our findings in Sec.~\ref{sec:analysis}, we seek an efficient linear alternative to the softmax which retains its spiky and monotonic properties. 
We first consider a simple potential approach: approximating the exponential in softmax by a low-degree Taylor polynomial~\citep{keles2023computational, Banerjee2020ExploringAT}. 

While in general, a high-quality approximation to the softmax should retain its spiky, monotonic, and performant properties, we ground our investigation with two potential caveats for the Taylor polynomial. First, recall that feature maps for $p$-degree polynomial approximations can be computed in $\gO(nd^p)$ time and space for every query and key vector. 
Thus, while this is indeed subquadratic in sequence length, the question remains whether we can set $p$ low enough to make the computation feasible while approximating $\exp$ reasonably.
Second, as a general property of polynomials, the 
Taylor approximation only tracks its original function with low error in bounded regimes. 

\textbf{Setup}. To test the Taylor approximation, we use the second-degree $\exp$ approximation, and evaluate on the prior train-from-scratch and finetuned-conversion settings (Sec.~\ref{sec:analysis_experiments}).
We implement the feature map as $\exp(\bm{q}^\top \bm{k} ) \approx \phi_\text{taylor}(\bm{q})^\top \phi_\text{taylor}(\bm{k})$, where $\phi_\text{taylor} (\bm{x})$ projects a $d$-dimensional query or key to $\gO(d^2)$-dimensional features
{\small
\(
\phi_\text{taylor}(\bm{x}) = \Big[1, x_1, \ldots, x_d, \Big] \cup \Big[x_i\cdot x_j \;| \; i,j \in [d] \Big]
\).
}
\textbf{Positive results.} We find that the 2nd-degree Taylor approximation retains 
both the spikiness and monotonic properties (Fig.~\ref{fig:taylor-swift}), and this corresponds to (near)-matching softmax attention performance (Table~\ref{table:tradeoff}).
We also note that here, the BERT query-key dot products are bounded in regimes where the second-order Taylor series $\exp$ approximation maintains monotonicity (Fig.~\ref{fig:taylor-swift}). This suggests we can enable expressive linear attentions for  training from scratch and finetuned-conversion.

\textbf{Caveats}. Unfortunately, the 2nd-degree Taylor approximation is not efficient. Even with $p=2$, the feature map dimension is now $d' = 1 + d + d^2$, resulting in $\gO(nd^3)$ attention complexity. As summarized in Table~\ref{table:tradeoff}, this introduces an efficiency-effectiveness trade-off among functional attention approximations. 
Thus, the question remains whether we can recover the expressivity and modeling quality of softmax while achieving similar $\gO(nd^2)$ scaling of past linear attentions.

\begin{table}[b!]
    \centering
    \begin{minipage}{0.55\textwidth}
        \centering
\resizebox{\linewidth}{!}{
\begin{tabular}{@{}lccccc@{}}
\toprule
Method     & Complexity       & Spiky? & \begin{tabular}[c]{@{}c@{}}Mono-\\ tonic?\end{tabular} & \begin{tabular}[c]{@{}c@{}}Train-from-\\ scratch (acc)\end{tabular} & \begin{tabular}[c]{@{}c@{}}BERT-FT\\ (MC)\end{tabular} \\ \midrule
Softmax    & $\gO(n^2d)$      & \cmark & \cmark                                                 & 100.0                                                               & 58.8                                                   \\ \midrule
1 + ELU    & $\bm{\gO(nd^2)}$ & \xmark & \xmark                                                 & 17.0                                                                & 28.1                                                   \\
Performer  & $\gO(nd'^2)$     & \xmark & \xmark                                                 & 17.0                                                                & 24.7                                                   \\
CosFormer  & $\bm{\gO(nd^2)}$ & \xmark & \xmark                                                 & 17.0                                                                & 39.9                                                   \\
Taylor Exp & $\gO(nd^3)$      & \cmark & \cmark                                                 & \textbf{100.0}                                                      & \textbf{58.4}                                          \\ \bottomrule
\end{tabular}
}
    \caption{Summary of feature maps compared to softmax, exhibiting an efficiency vs. expressivity tradeoff. 
    }
\label{table:tradeoff}

    \end{minipage} \hfill
    \begin{minipage}{0.425\textwidth}
  \centering
 \includegraphics[width=1\linewidth]{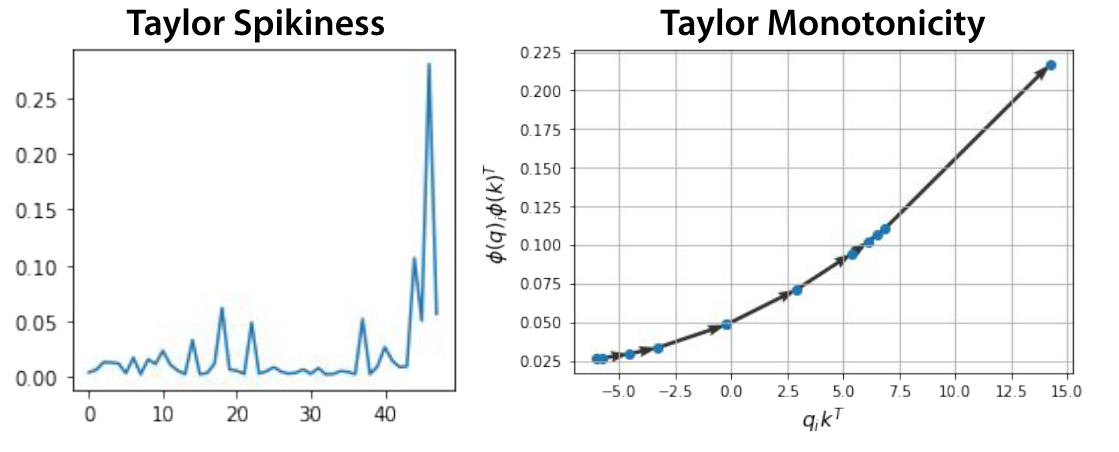}
\vspace{-0.5cm}
\captionof{figure}{Taylor approximation recovers spikiness and monotonicity}
\label{fig:taylor-swift}
    \end{minipage}\hfill
    \vspace{-0.5cm}
\end{table}

\subsection{Learnable Linear Attentions for Mimicking Softmax}
\label{sec:hedgehog_method}

Our key insight is that rather than rely on fixed functional form that captures our spiky and monotonic properties, we can learn linear attention feature maps that do so. For each attention block, we propose feature maps as trainable single-layer MLPs, which is similar to prior work~\citep{kasai-etal-2021-finetuning} and acts similarly to an adapter~\citep{houlsby2019parameter} inserted after the query and key projections in Transformer attention layers (Fig.~\ref{fig:pull_fig}).
However, unlike prior work, we explicitly train these feature maps such that the attention layers mimic the properties of softmax attention. \update{We describe these two core components below, and validate these design choices in Sec.~\ref{sec:hedgehog_benchmarking}}.

\paragraph{Spiky MLP feature map.} 
\label{sec:lk_arch}
Recall the kernel based linear attention paradigm from Sec.~\ref{sec:preliminaries}, where a feature map $\phi : \mathbb{R}^d \mapsto \mathbb{R}^{d'}$ is applied to both queries and keys to compute causal self-attention outputs using~\eqref{eq:linear_attention}.
%
However, unlike prior work that sticks to a pre-specified function as a feature map, we make the feature map a trainable MLP. 
%
%
In particular, for the single-head attention setting, we compute $\phi_{\text{mlp}}(\bm{q}_i)^\top\phi_{\text{mlp}}(\bm{k}_j)$ with a simple one-layer MLP as 
 $   \phi_{\text{mlp}}(\bm{x}) = \Phi(\bm{W}^\top \bm{x} + \bm{b})$ 
where the matrix $\bm{W} \in \mathbb{R}^{d \times d'}$ and the bias $\bm{b} \in \mathbb{R}^{d'}$ are learned, and $\Phi$ is an activation function. To induce spikiness, we set $\Phi$ as the element-wise exponential function studied in Sec.~\ref{sec:analysis_experiments}, resulting in
{\small
\begin{equation}
    \phi_{\text{mlp}}(\bm{x}) = \Big[\exp(\bm{w}_1^\top \bm{x} + \bm{b}), \ldots, \exp(\bm{w}_d^\top \bm{x}  + \bm{b})\Big]
    \label{eq:expanded_hh}
\end{equation}
}



\paragraph{Attention weight distillation loss.}
\label{sec:lk_loss}
To learn a softmax approximation, we train $\phi_{\text{mlp}}$ to minimize the cross-entropy loss between the computed linear attention weights and those that would have been computed via softmax attention. For query $\bm{q}_i$ and keys $\{\bm{k}_j\}_1^n$, we compute the sample losses as
{\small
\begin{equation}
    \mathcal{L}_i = -\sum_{j = 1}^{i} 
\frac{\exp(\bm{q}_i^\top\bm{k}_j)}{\sum_{m=1}^{i} \exp(\bm{q}_i^\top\bm{k}_m)} \log
\frac{\phi_{\text{mlp}}(\bm{q}_i)^\top\phi_{\text{mlp}}(\bm{k}_j)}{\sum_{m=1}^{i} \phi_{\text{mlp}}(\bm{q}_i)^\top\phi_{\text{mlp}}(\bm{k}_j)}
\label{eq:distillation_loss}
\end{equation}
}

For training \name{} attentions in multi-layer and multi-head attention Transformers, we apply a separate MLP to each head and each layer, and use the same $\phi_\text{mlp}$ for the queries and keys. We include further implementation details and pseudocode in Appendix~\ref{app:hh_implement_deets}.

\begin{table}[b]
    \centering
    \begin{minipage}{0.525\textwidth}
        \centering
\begin{tabular}{@{}lccc@{}}
\toprule
Method     & Complexity          & AR & BERT-FT \\ \midrule
Softmax    & $\mathcal{O}(n^2d)$ & 100.0              & 58.8          \\
Taylor Exp & $\mathcal{O}(nd^3)$ & 100.0              & 58.4          \\
Hedgehog   & $\bm{\mathcal{O}(nd^2)}$ & \textbf{100.0}              & \textbf{59.2}          \\ \bottomrule
\end{tabular}
    \caption{\update{\name{} matches performance on associative recall (AR) and BERT-finetuned conversion (BERT-FT) with prior best approaches, while achieving better time and space complexity.}}
        \label{tab:assoc_recall_hh}
    \end{minipage}\hfill
    \begin{minipage}{0.45\textwidth}
        \centering
        \includegraphics[width=\textwidth]{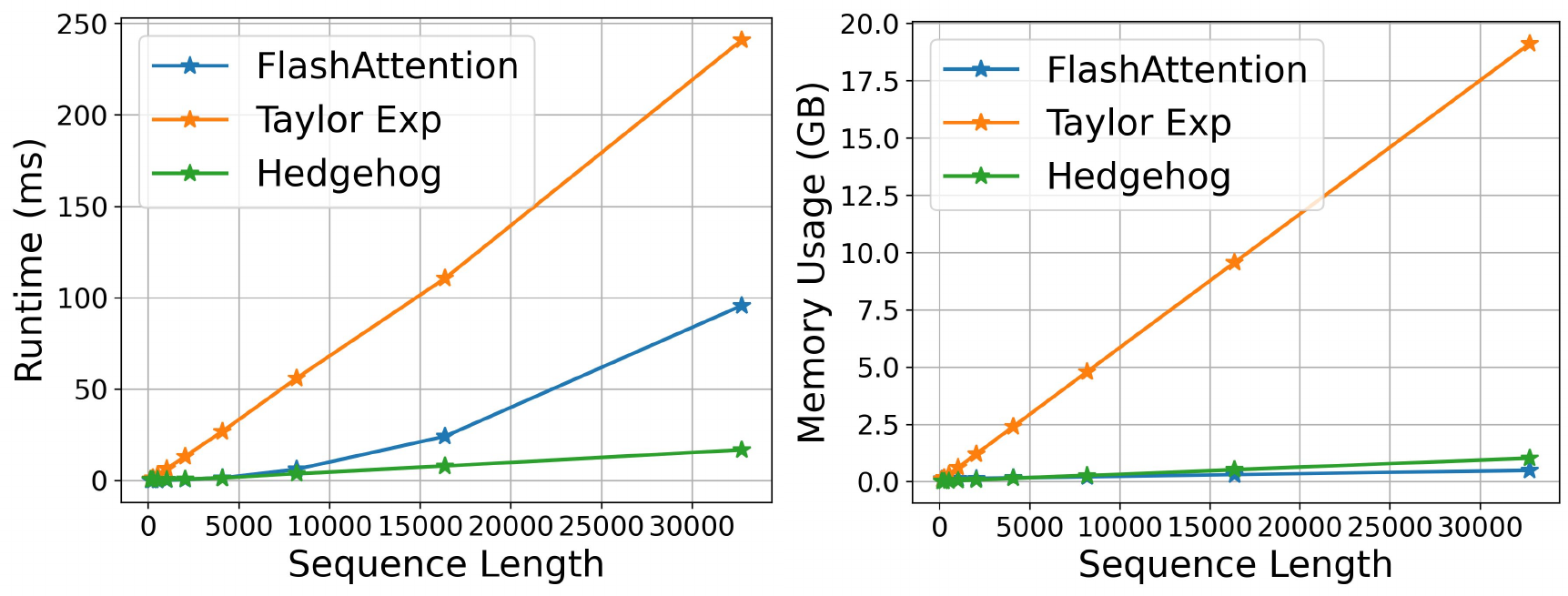}
        \vspace{-0.5cm}
        \captionof{figure}{\name{} linear scaling in wall-clock time (left) and memory (right). Unlike the Taylor approx., \name{} inference gets real-world gains over FlashAttention.}
        \label{fig:scaling}
    \end{minipage}
\vspace{-0.35cm}
\end{table}

\section{Experiments}
\label{sec:results}

In experiments, we evaluate whether \name{} recovers softmax attention expressivity while retaining linear attention efficiency (Sec.~\ref{sec:hedgehog_benchmarking}), and how this improves modeling quality in training-from-scratch (Sec.~\ref{sec:results_from_scratch}), finetuned-conversion (Sec.~\ref{sec:results_from_finetuned}), and pretrained-conversion regimes (Sec.~\ref{sec:results_from_pretrained}).

\subsection{Benchmarking \name{} for Expressivity and Efficiency}
\label{sec:hedgehog_benchmarking}
Before evaluating \name{} on downstream tasks, we aim to validate \name{}'s design choices for efficiency and expressivity. We address: (1) Do \name{}'s spiky feature map and distillation loss recover the spiky and monotonic properties of softmax attention on the prior associative recall and BERT CoLA tasks? (2) Does \name {} achieve improved efficiency over softmax attention? \update{(3) For conversion, do the learned attention weights actually match those of ``ground-truth'' softmax attention? Once learned, does this transfer to longer contexts and different tasks?}


\textbf{Recovering softmax spiky and monotonic properties}. We test \name{} in the same train-from-scratch associative recall (AR) and finetuned-conversion of BERT on CoLA settings in Sec.~\ref{sec:analysis_experiments}. For training-from-scratch on AR, we do not use the distillation loss, and train the model end-to-end with next-token-prediction after inserting the learnable MLPs. In Table.~\ref{tab:assoc_recall_hh}, we find that \name{} achieves both favorable complexity and modeling for train-from-scratch and finetuned-conversion. This corresponds respectively with the spiky (Fig.~\ref{fig:entropy_viz}) and monotonic (Fig.~\ref{fig:monotonic_viz}) properties noted prior.

\textbf{Recovering linear attention efficiency.} We next find \name{}'s $\gO(nd^2)$ scaling in compute and memory can lead to real-world efficiency gains. We benchmark inference in wall-clock time and memory usage for one attention layer with 12 heads and head dimension = 64 on sequences up to $n$ = 32K tokens long (Fig.~\ref{fig:scaling}). \name{} achieves near 6x faster inference and similar memory to FlashAttention~\citep{dao2022flashattention} (linear in memory but quadratic in time). Meanwhile, the Taylor approximation, while $\mathcal{O}(n)$, gets significantly larger memory and slower speed due to the extra $d$.

\begin{figure}[t] 
    \centering
    \includegraphics[width=1\textwidth]{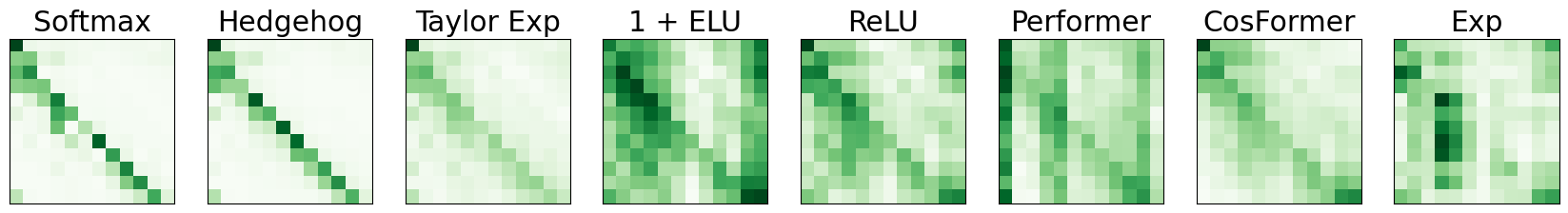}
    \vspace{-0.5cm}
    \caption{Compared to prior linear attentions, trained \name{} layers (\textbf{2nd left}) produce attention weights closely tracking softmax (\textbf{left}), with greater fidelity with both components (vs. Fig.~\ref{fig:hh_ablate}).}
    \label{fig:attention_matching}
    \vspace{-0.25cm}
\end{figure}

\textbf{Recovering softmax attention weights}. We next study the \emph{combination} of \name{}'s feature map and distillation loss for matching softmax attention weights. Beyond recovering the spiky and monotonic properties, learning to exactly match the weights can be particularly effective for converting or ``distilling'' pretrained quadratic Transformers into linear variants.
For evaluation, we visualize the attention weights for different linear attentions in our BERT-FT CoLA setting (Fig.~\ref{fig:attention_matching}). We find \name{} recovers linear attention weights that match softmax's with much higher fidelity.

\update{To further understand the contribution of \name{}'s (1) spiky MLP and (2) distillation loss in Sec.~\ref{sec:hedgehog_method}, we visualize ablated attention weights by (1) using the distillation loss with the ReLU feature map used in Transformer-to-RNN (T2R-HH) (\cite{kasai-etal-2021-finetuning}), and (2) using untrained MLPs, replacing the trainable weights with an identity function (HH No Train). We find that distillation training is necessary to recover attention weights, and that the spiky MLP is also helpful for matching attentions (later supported by improved Transformer conversion in Sec.~\ref{sec:results_from_finetuned}).}

\begin{table}[t]
    \centering
    \begin{minipage}{0.25\textwidth}
        \centering
        \includegraphics[width=\textwidth]{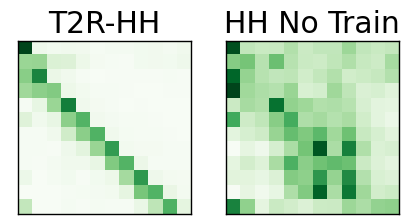}
        \vspace{-0.25cm}
        \captionof{figure}{Hedgehog ablated attention weights.}
        \label{fig:hh_ablate}
    \end{minipage}\hfill
    \begin{minipage}{0.725\textwidth}
        \centering
\resizebox{\linewidth}{!}{
\begin{tabular}{@{}lccccccc@{}}
\toprule
Dataset & \begin{tabular}[c]{@{}c@{}}HH \\ (CoLA)\end{tabular} & \begin{tabular}[c]{@{}c@{}}HH \\ (WT-103)\end{tabular} & \begin{tabular}[c]{@{}c@{}}T2R-HH \\ (CoLA)\end{tabular} & \begin{tabular}[c]{@{}c@{}}HH \\ (No Train)\end{tabular} & \begin{tabular}[c]{@{}c@{}}1 +\\ ELU\end{tabular} & Performer & CosFormer \\ \midrule
CoLA             & \textbf{0.172}                                       & 0.352                                                  & \underline{0.191}                                              & 0.694                                                    & 1.223                                             & 1.293     & 1.196     \\
MRPC             & \underline{0.663}                                          & \textbf{0.485}                                         & 1.128                                                    & 1.250                                                     & 2.165                                             & 2.234     & 1.982     \\
MNLI             & \textbf{0.345}                                       & \underline{0.382}                                            & 0.613                                                    & 0.851                                                    & 1.51                                              & 1.581     & 1.338     \\
QNLI             & \underline{0.671}                                          & \textbf{0.444}                                         & 1.126                                                    & 1.139                                                    & 1.968                                             & 2.069     & 1.817     \\ \bottomrule
\end{tabular}
}
    \caption{\update{We find \name{} feature maps trained via distillation on CoLA or WikiText-103 generalize to new GLUE data, better matching softmax than prior linear attentions or ablations (reporting KL div.).}}
        \label{table:generalize_kl}
    \end{minipage}
    \vspace{-0.25cm}
\end{table}

\begin{table}[t!]
    \centering
    \begin{minipage}{0.575\textwidth}
        \centering
        \includegraphics[width=\textwidth]{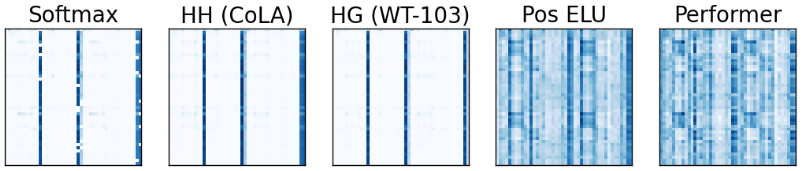}
        \vspace{-0.5cm}
        \captionof{figure}{\name{} trained on CoLA and WT-103 recover softmax attentions on MRPC data.}
        \label{fig:generalize_data_viz}
    \end{minipage}\hfill
    \begin{minipage}{0.4\textwidth}
        \centering
\resizebox{\linewidth}{!}{
\begin{tabular}{@{}lccccc@{}}
\toprule
Seq. Len & 256      & 1024  & 2048  & 4096  \\ \midrule
CoLA KL  & 0.182  & 0.187 & 0.190 & 0.181 \\ \bottomrule
\end{tabular}
}
    \caption{\name{} attention maintains fidelity with softmax attention over context lengths for BERT-FT on CoLA.}
        \label{table:generalize_context_len}
    \end{minipage}
    \vspace{-0.5cm}
\end{table}

\textbf{Generalization to new data and longer contexts.} \update{Finally, we investigate the generality of learned \name{} feature maps. We show \name{} attentions learned over specific data and context lengths can still better match softmax attention weights for new data and sequence lengths than prior linear attentions. We distill attentions for BERT models using CoLA or WikiText-103 (WT-103) samples, and report attention weights compared to softmax attention on three other GLUE tasks: qualitatively (Fig.~\ref{fig:generalize_data_viz}) and quantitatively via KL divergence w.r.t. the ``ground-truth'' softmax weights (Table~\ref{table:generalize_kl}). We include additional visualizations and comparisons Appendix~\ref{table:generalize_context_len}.}

\update{In Table~\ref{fig:generalize_data_viz}, we further show that \name{} attention matching remains consistent across longer contexts. Post-distillation on CoLA samples, we concatenate CoLA samples into sequences 256 to 4096 tokens long (up to 8x the default 512 context length). We then compute attention weights using softmax and learned \name{} feature maps, and find that their KL divergence remains consistent.} 

\subsection{Learning Sequence Modeling From Scratch}
\label{sec:results_from_scratch}
We evaluate \name{} Transformers trained from scratch on the popular LRA sequence classification and WikiText-103 language modeling benchmarks. 
\update{For training from scratch, we initialize MLPs as identity matrices for \name{} feature maps}, and train the entire models end-to-end with the task-specific loss. We find \name{} achieves best average accuracy for both tasks among linear attentions (Table~\ref{table:LRA},~\ref{table:scratch_wt103}). \update{For LRA, while non-Transformer models are now state-of-the-art~\citep{gu2021efficiently}, our work focuses on approximating attention, so we compare with competitive subquadratic Transformers. We adopt the same hyperparameter settings as the official benchmark~\citep{tay2021long}}. On WikiText-103, we adopt the setting in~\cite{fu2023hungry}, evaluating a 125M decoder-only Transformer on perplexity over 1024 tokens. \name{} significantly closes the gap by up to 6 PPL. 
{\footnotesize
\begin{table}[t]
\begin{center}
\resizebox{0.75\linewidth}{!}{
\begin{tabular}{@{}lcccccc@{}}
\toprule
Model           & ListOps        & Text           & Retrieval      & Image          & Pathfinder     & Average            \\ \midrule
Transformer     & 36.37          & 64.27          & 57.46          & 42.44          & 71.40          & 54.39          \\ \midrule
Local Att       & 15.82          & 52.98          & 53.39          & 41.46          & 66.63          & 46.06          \\
Linear Trans.   & 16.13          & \textbf{65.90}          & 53.09          & {42.34}    & \underline{75.30}    & 50.55          \\
Reformer        & 37.27          & 56.10          & 53.40          & 38.07          & 68.50          & 50.67          \\
Sparse Trans.   & 17.07          & 63.58          & 59.59          & 44.24          & 71.71          & 51.24          \\
Sinkhorn Trans. & 33.67          & 61.20          & 53.83          & 41.23          & 67.45          & 51.29          \\
Linformer       & 35.70          & 53.94          & 52.27          & 38.56          & 76.34          & 51.36          \\
Performer       & 18.01          & 65.40          & 53.82          & \underline{42.77} & \textbf{77.05} & 51.41          \\
Synthesizer     & 36.99          & 61.68          & 54.67          & 41.61          & 69.45          & 52.88          \\
Longformer      & 35.63          & 62.85          & 56.89          & 42.22          & 69.71          & 53.46          \\
BigBird         & 36.05          & 64.02          & 59.29          & 40.83          & 74.87          & 55.01          \\
Nyströmformer$^\dagger$   & 37.15          & \underline{65.52}          & 79.56          & 41.58          & 70.94          & 58.95          \\
cosFormer$^\dagger$       & \underline{37.90}    & 63.41          & 61.36          & \textbf{43.17}          & 70.33          & 55.23          \\
Skyformer$^\dagger$       & \textbf{39.25} & 64.70  & \underline{82.06}    & 40.77          & 70.73          & \underline{59.50}    \\  \midrule
Hedgehog        & 37.15          & {64.60}    & \textbf{82.24} & 40.15          & 74.16          & \textbf{59.66} \\ \bottomrule
\end{tabular}
}
\end{center}
\caption{Training-from-scratch on LRA. \name{} achieves best avg. acc. (\%) across most competitive Transformers \update{(full results in Table~\ref{table:LRA_full}, trends hold)}. $^\dagger$ indicates results reported from original works. All others reported from the official LRA benchmark~\citep{tay2021long}. 
\textbf{Best}, \underline{2nd-best}.
} 
\label{table:LRA}
\end{table}
}

\begin{table}[t]
\begin{center}
\resizebox{0.75\linewidth}{!}{
\begin{tabular}{@{}lcccccc@{}}
\toprule
Model      & Transformer   & Performer & Reformer & AFT  & (1 + ELU) & Hedgehog   \\ \midrule
Perplexity & \textbf{18.6} & 26.8      & 25.6     & 28.2 & 25.6               & \underline{20.8} \\ \bottomrule
\end{tabular}
}
\end{center}
\vspace{-2mm}
\caption{Training-from-scratch on WikiText-103. Among 125M decoder-only models, \name{} significantly closes the gap between standard Transformers and prior linear attention maps by 68.6\%.
}
\label{table:scratch_wt103}
\vspace{-0.5cm}
\end{table}

\subsection{Finetuned Conversion of Quadratic to Linear Transformers}
\label{sec:results_from_finetuned}
For the finetuned Transformer conversion regime, we evaluate performance recovery for BERT-base models finetuned on GLUE, \update{and ViT-B/16 models trained on ImageNet-1K}.
For both settings, we first swap attentions and train via our distillation loss (Sec.~\ref{sec:lk_loss}).
We then finetune the converted BERT models on their original tasks as in Transformer-to-RNN (T2R)~\citep{kasai-etal-2021-finetuning}.

For BERT, we compare \name{} to T2R in Table~\ref{table:ft_conversion}, and find that in contrast, \name{} conversion recovers near-100\% of the original softmax attention performance. 
To further test \name{}'s feature map and attention distillation, we also compare against an ablation that trains the T2R feature map with our distillation loss (T2R-HH). We find that training to mimic softmax attentions boosts performance of T2R, suggesting that attention weight distillation may be a general step to improving linear attention feature maps. However, \name{}'s exponential still leads to superior performance. \update{We find similar results for ViT-B/16, suggesting \name{} can also apply to other modalities.}

\subsection{Pretrained Conversion for Subquadratic Task Transfer}
\label{sec:results_from_pretrained}

We finally evaluate \name{} for converting pretrained Transformers into linear Transformers.
\update{We consider two settings:} (1) To benchmark \name{} and the pretrained-conversion regime for subquadratic sequence modeling, we use the same WT-103 evaluation in Sec.~\ref{sec:results_from_scratch} for converting 125M-parameter GPT-2. \update{(2) As an early application for \name{} on larger models, we convert Llama-2 7B~\citep{touvron2023llama} before finetuning with low-rank adapters (LoRA)~\citep{hu2021lora} on SAMSum summarization~\citep{gliwa-etal-2019-samsum}.} 
We include further training details in Appendix.~\ref{app:train_pretrained_deets}. 

To most directly measure pretrained-conversion quality, for both settings we compare against T2R. For GPT-2, we find \name{} both outperforms T2R, and further outperforms modern subquadratic sequence models such as H3~\citep{fu2023hungry} and Hyena~\citep{poli2023hyena} (Table~\ref{table:pt_wt103}). 
Although not directly comparable due to pretraining, we also compare with  zero-shot and finetuned GPT-2 for reference. 
While \name{} is 1 PPL off the fully quadratic finetuned GPT-2, it significantly improves over zero-shot while being linear to train. 
\update{We finally apply \name{} for Llama-2 conversion, where \name{} enables linear attention Llamas that train via LoRA (see Appendix~\ref{app:llama_gen} for sample generations). 
}

\begin{table}[h]
    \centering
    \begin{minipage}{0.75\textwidth}
        \centering
\resizebox{\linewidth}{!}{
\begin{tabular}{@{}lccccccccc@{}}
\toprule
Method            & CoLA     & SST2          & MRPC     & STS-B & QQP & MNLI & QNLI          & RTE                  & (\%) Recover \\ \midrule
BERT-FT           & 58.8          & 93.2          & 90.2          & 88.8            & 91.0    & 84.7 & 91.3          & 68.2                    & 100.0          \\ \midrule
T2R               & 43.6          & 87.7          & 83.0          & 78.6            & 86.7    &   78.9   & 84.6          & 54.1                   & 88.9           \\
T2R-HH & 56.9          & 90.9          & 89.1          & 77.7            & 90.0    & 77.4 & 84.5          & 56.3                    & 93.5           \\
Hedgehog & \textbf{59.2} & \textbf{92.6} & \textbf{90.1} & \textbf{87.4}   & \textbf{91.0}    & \textbf{82.6} & \textbf{89.6} & \textbf{69.3}  & \textbf{99.3}  \\ \bottomrule
\end{tabular}
}
\vspace{-2mm}
\caption{Finetuned-conversion evaluation. \name{} recovers 99.3\% of original finetuned BERT (BERT-FT) GLUE performance. 
} 
\label{table:ft_conversion}
    \end{minipage}\hfill
    \begin{minipage}{0.225\textwidth}
        \centering
        \resizebox{\linewidth}{!}{
        \begin{tabular}{@{}lc@{}}
\toprule
      Top-1   &  Acc. \% \\ \midrule
ViT-B/16 & 80.3                                                       \\ \midrule
T2R-HH   & 77.0                                                       \\
Hedgehog & \textbf{79.5}                                                       \\ \bottomrule
\end{tabular}
}
\vspace{-0.25cm}
\caption{\update{\name{} achieve 99\% ViT acc.}} 
\label{table:ft_conversion_vit}
    \end{minipage}
\vspace{-0.35cm}
\end{table}

\begin{table}[h!]
    \centering
    \begin{minipage}{0.67\textwidth}
        \centering
\resizebox{\linewidth}{!}{
\begin{tabular}{@{}lcccccc@{}}
\toprule
Method           & GPT-2  & \update{GPT-2 FT} & Hybrid H3  & Hyena  &  T2R-GPT-2  & HH-GPT-2  \\ \midrule
PPL & 28.0 & \update{15.8}      & 18.5      & 18.5 & 19.4  & \textbf{16.7}     \\ \bottomrule
\end{tabular}
}
\caption{Pretrained-conversion for 125M GPT-2 on WT-103 lang. modeling. \update{While finetuned GPT-2 gets lowest PPL, among \emph{subquadratic} models \name{} significantly outperforms by 1.8 PPL. }}
\label{table:pt_wt103}
    \end{minipage}\hfill
    \begin{minipage}{0.3\textwidth}
        \centering
        \resizebox{\linewidth}{!}{
\begin{tabular}{@{}lc@{}}
\toprule
Llama-2         & R1 / R2 / RL       \\ \midrule
Softmax (Zero-shot) & 19.3 / 6.8 / 14.9 \\ 
Softmax (LoRA) & 51.1 / 27.6 / 43.5 \\ \midrule
T2R (LoRA)     & 2.8 / 0.0 / 2.6      \\
Hedgehog (LoRA) & \textbf{47.4 / 23.4 / 39.1} \\ \bottomrule
\end{tabular}
}
\vspace{-0.25cm}
\caption{\update{\name{} Llama-2 conversion (ROUGE).}}
\label{table:pt_llama}
    \end{minipage}
\vspace{-0.5cm}
\end{table}

\vspace{-0.25cm}
\section{Conclusion}
\label{sec:conclusion}
\vspace{-0.25cm}
We present \name{}, a learnable linear attention to mimic softmax attention. This enables training linear attention models from scratch and \emph{converting} existing Transformers into linear attention variants. 
To motivate \name{} we study why prior linear attentions underperform softmax attention, and identify two missing properties: (1) the ability to capture low entropy or spiky attention maps and (2) to be monotonic with respect to the underlying query-key dot products. We find training to match softmax attentions results in recovering many of its expressive properties, and that \name{} leads to competitive performance with softmax-based attention in training from scratch, finetuned-conversion, and pretrained conversion regimes. 

\section*{Acknowledgements}
We thank Armin Thomas, Gordon Downs, Krista Opsahl-Ong, Pun Waiwitlikhit, Schwinn Saereesitthipitak, Dan Fu, Simran Arora, Sabri Eyuboglu, and Tri Dao for helpful discussions on linear attention and paper feedback, and Dan Fu for prior versions of the pseudocode formatting in the appendix.

We gratefully acknowledge the support of NIH under No. U54EB020405 (Mobilize), NSF under Nos. CCF1763315 (Beyond Sparsity), CCF1563078 (Volume to Velocity), and 1937301 (RTML); US DEVCOM ARL under No. W911NF-21-2-0251 (Interactive Human-AI Teaming); ONR under No. N000141712266 (Unifying Weak Supervision); ONR N00014-20-1-2480: Understanding and Applying Non-Euclidean Geometry in Machine Learning; N000142012275 (NEPTUNE); NXP, Xilinx, LETI-CEA, Intel, IBM, Microsoft, NEC, Toshiba, TSMC, ARM, Hitachi, BASF, Accenture, Ericsson, Qualcomm, Analog Devices, Google Cloud, Salesforce, Total, the HAI-GCP Cloud Credits for Research program,  the Stanford Data Science Initiative (SDSI), and members of the Stanford DAWN project: Facebook, Google, and VMWare. The U.S. Government is authorized to reproduce and distribute reprints for Governmental purposes notwithstanding any copyright notation thereon. Any opinions, findings, and conclusions or recommendations expressed in this material are those of the authors and do not necessarily reflect the views, policies, or endorsements, either expressed or implied, of NIH, ONR, or the U.S. Government.

\bibliography{main.bib}

\begin{thebibliography}{41}
\providecommand{\natexlab}[1]{#1}
\providecommand{\url}[1]{\texttt{#1}}
\expandafter\ifx\csname urlstyle\endcsname\relax
  \providecommand{\doi}[1]{doi: #1}\else
  \providecommand{\doi}{doi: \begingroup \urlstyle{rm}\Url}\fi

\bibitem[Alman \& Song(2023)Alman and Song]{alman2023fast}
Josh Alman and Zhao Song.
\newblock Fast attention requires bounded entries.
\newblock \emph{arXiv preprint arXiv:2302.13214}, 2023.

\bibitem[Ba et~al.(2016)Ba, Hinton, Mnih, Leibo, and Ionescu]{ba2016using}
Jimmy Ba, Geoffrey~E Hinton, Volodymyr Mnih, Joel~Z Leibo, and Catalin Ionescu.
\newblock Using fast weights to attend to the recent past.
\newblock \emph{Advances in neural information processing systems}, 29, 2016.

\bibitem[Bahdanau et~al.(2014)Bahdanau, Cho, and Bengio]{bahdanau2014neural}
Dzmitry Bahdanau, Kyunghyun Cho, and Yoshua Bengio.
\newblock Neural machine translation by jointly learning to align and translate.
\newblock \emph{arXiv preprint arXiv:1409.0473}, 2014.

\bibitem[Banerjee et~al.(2020)Banerjee, C., Gupta, Vyas, H., and Mishra]{Banerjee2020ExploringAT}
Kunal Banerjee, Vishak C., Rishi~Raj Gupta, Kartik Vyas, Anushree H., and Biswajit Mishra.
\newblock Exploring alternatives to softmax function.
\newblock \emph{ArXiv}, abs/2011.11538, 2020.
\newblock URL \url{https://api.semanticscholar.org/CorpusID:227127574}.

\bibitem[Biderman et~al.(2023)Biderman, Schoelkopf, Anthony, Bradley, O’Brien, Hallahan, Khan, Purohit, Prashanth, Raff, et~al.]{biderman2023pythia}
Stella Biderman, Hailey Schoelkopf, Quentin~Gregory Anthony, Herbie Bradley, Kyle O’Brien, Eric Hallahan, Mohammad~Aflah Khan, Shivanshu Purohit, USVSN~Sai Prashanth, Edward Raff, et~al.
\newblock Pythia: A suite for analyzing large language models across training and scaling.
\newblock In \emph{International Conference on Machine Learning}, pp.\  2397--2430. PMLR, 2023.

\bibitem[Caron et~al.(2021)Caron, Touvron, Misra, J{\'e}gou, Mairal, Bojanowski, and Joulin]{caron2021emerging}
Mathilde Caron, Hugo Touvron, Ishan Misra, Herv{\'e} J{\'e}gou, Julien Mairal, Piotr Bojanowski, and Armand Joulin.
\newblock Emerging properties in self-supervised vision transformers.
\newblock In \emph{Proceedings of the IEEE/CVF international conference on computer vision}, pp.\  9650--9660, 2021.

\bibitem[Chen et~al.(2023)Chen, Wong, Chen, and Tian]{chen2023extending}
Shouyuan Chen, Sherman Wong, Liangjian Chen, and Yuandong Tian.
\newblock Extending context window of large language models via positional interpolation.
\newblock \emph{arXiv preprint arXiv:2306.15595}, 2023.

\bibitem[Chen et~al.(2021)Chen, Zeng, Ji, and Yang]{chen2021skyformer}
Yifan Chen, Qi~Zeng, Heng Ji, and Yun Yang.
\newblock Skyformer: Remodel self-attention with gaussian kernel and nystr{\textbackslash}''om method.
\newblock In A.~Beygelzimer, Y.~Dauphin, P.~Liang, and J.~Wortman Vaughan (eds.), \emph{Advances in Neural Information Processing Systems}, 2021.
\newblock URL \url{https://openreview.net/forum?id=pZCYG7gjkKz}.

\bibitem[Choromanski et~al.(2020)Choromanski, Likhosherstov, Dohan, Song, Gane, Sarlos, Hawkins, Davis, Mohiuddin, Kaiser, et~al.]{choromanski2020rethinking}
Krzysztof Choromanski, Valerii Likhosherstov, David Dohan, Xingyou Song, Andreea Gane, Tamas Sarlos, Peter Hawkins, Jared Davis, Afroz Mohiuddin, Lukasz Kaiser, et~al.
\newblock Rethinking attention with performers.
\newblock \emph{arXiv preprint arXiv:2009.14794}, 2020.

\bibitem[Choromanski et~al.(2021)Choromanski, Chen, Lin, Ma, Sehanobish, Jain, Ryoo, Varley, Zeng, Likhosherstov, et~al.]{choromanski2021hybrid}
Krzysztof Choromanski, Haoxian Chen, Han Lin, Yuanzhe Ma, Arijit Sehanobish, Deepali Jain, Michael~S Ryoo, Jake Varley, Andy Zeng, Valerii Likhosherstov, et~al.
\newblock Hybrid random features.
\newblock \emph{arXiv preprint arXiv:2110.04367}, 2021.

\bibitem[Dai et~al.(2019)Dai, Yang, Yang, Carbonell, Le, and Salakhutdinov]{dai-etal-2019-transformer}
Zihang Dai, Zhilin Yang, Yiming Yang, Jaime Carbonell, Quoc Le, and Ruslan Salakhutdinov.
\newblock Transformer-{XL}: Attentive language models beyond a fixed-length context.
\newblock In \emph{Proceedings of the 57th Annual Meeting of the Association for Computational Linguistics}, pp.\  2978--2988, Florence, Italy, July 2019. Association for Computational Linguistics.
\newblock \doi{10.18653/v1/P19-1285}.
\newblock URL \url{https://aclanthology.org/P19-1285}.

\bibitem[Dao et~al.(2022)Dao, Fu, Ermon, Rudra, and R{\'e}]{dao2022flashattention}
Tri Dao, Dan Fu, Stefano Ermon, Atri Rudra, and Christopher R{\'e}.
\newblock Flashattention: Fast and memory-efficient exact attention with io-awareness.
\newblock \emph{Advances in Neural Information Processing Systems}, 35:\penalty0 16344--16359, 2022.

\bibitem[Dettmers et~al.(2023)Dettmers, Pagnoni, Holtzman, and Zettlemoyer]{dettmers2023qlora}
Tim Dettmers, Artidoro Pagnoni, Ari Holtzman, and Luke Zettlemoyer.
\newblock Qlora: Efficient finetuning of quantized llms.
\newblock \emph{arXiv preprint arXiv:2305.14314}, 2023.

\bibitem[Devlin et~al.(2018)Devlin, Chang, Lee, and Toutanova]{devlin2018bert}
Jacob Devlin, Ming-Wei Chang, Kenton Lee, and Kristina Toutanova.
\newblock Bert: Pre-training of deep bidirectional transformers for language understanding.
\newblock \emph{arXiv preprint arXiv:1810.04805}, 2018.

\bibitem[Dosovitskiy et~al.(2020)Dosovitskiy, Beyer, Kolesnikov, Weissenborn, Zhai, Unterthiner, Dehghani, Minderer, Heigold, Gelly, et~al.]{dosovitskiy2020image}
Alexey Dosovitskiy, Lucas Beyer, Alexander Kolesnikov, Dirk Weissenborn, Xiaohua Zhai, Thomas Unterthiner, Mostafa Dehghani, Matthias Minderer, Georg Heigold, Sylvain Gelly, et~al.
\newblock An image is worth 16x16 words: Transformers for image recognition at scale.
\newblock \emph{arXiv preprint arXiv:2010.11929}, 2020.

\bibitem[Fu et~al.(2023)Fu, Dao, Saab, Thomas, Rudra, and Re]{fu2023hungry}
Daniel~Y Fu, Tri Dao, Khaled~Kamal Saab, Armin~W Thomas, Atri Rudra, and Christopher Re.
\newblock Hungry hungry hippos: Towards language modeling with state space models.
\newblock In \emph{The Eleventh International Conference on Learning Representations}, 2023.
\newblock URL \url{https://openreview.net/forum?id=COZDy0WYGg}.

\bibitem[Gliwa et~al.(2019)Gliwa, Mochol, Biesek, and Wawer]{gliwa-etal-2019-samsum}
Bogdan Gliwa, Iwona Mochol, Maciej Biesek, and Aleksander Wawer.
\newblock {SAMS}um corpus: A human-annotated dialogue dataset for abstractive summarization.
\newblock In Lu~Wang, Jackie Chi~Kit Cheung, Giuseppe Carenini, and Fei Liu (eds.), \emph{Proceedings of the 2nd Workshop on New Frontiers in Summarization}, pp.\  70--79, Hong Kong, China, November 2019. Association for Computational Linguistics.
\newblock \doi{10.18653/v1/D19-5409}.
\newblock URL \url{https://aclanthology.org/D19-5409}.

\bibitem[Gu et~al.(2021)Gu, Goel, and R{\'e}]{gu2021efficiently}
Albert Gu, Karan Goel, and Christopher R{\'e}.
\newblock Efficiently modeling long sequences with structured state spaces.
\newblock \emph{arXiv preprint arXiv:2111.00396}, 2021.

\bibitem[Houlsby et~al.(2019)Houlsby, Giurgiu, Jastrzebski, Morrone, De~Laroussilhe, Gesmundo, Attariyan, and Gelly]{houlsby2019parameter}
Neil Houlsby, Andrei Giurgiu, Stanislaw Jastrzebski, Bruna Morrone, Quentin De~Laroussilhe, Andrea Gesmundo, Mona Attariyan, and Sylvain Gelly.
\newblock Parameter-efficient transfer learning for nlp.
\newblock In \emph{International Conference on Machine Learning}, pp.\  2790--2799. PMLR, 2019.

\bibitem[Hu et~al.(2021)Hu, Shen, Wallis, Allen-Zhu, Li, Wang, Wang, and Chen]{hu2021lora}
Edward~J Hu, Yelong Shen, Phillip Wallis, Zeyuan Allen-Zhu, Yuanzhi Li, Shean Wang, Lu~Wang, and Weizhu Chen.
\newblock Lora: Low-rank adaptation of large language models.
\newblock \emph{arXiv preprint arXiv:2106.09685}, 2021.

\bibitem[Irie et~al.(2021)Irie, Schlag, Csord{\'a}s, and Schmidhuber]{irie2021going}
Kazuki Irie, Imanol Schlag, R{\'o}bert Csord{\'a}s, and J{\"u}rgen Schmidhuber.
\newblock Going beyond linear transformers with recurrent fast weight programmers.
\newblock In A.~Beygelzimer, Y.~Dauphin, P.~Liang, and J.~Wortman Vaughan (eds.), \emph{Advances in Neural Information Processing Systems}, 2021.
\newblock URL \url{https://openreview.net/forum?id=ot2ORiBqTa1}.

\bibitem[Kasai et~al.(2021)Kasai, Peng, Zhang, Yogatama, Ilharco, Pappas, Mao, Chen, and Smith]{kasai-etal-2021-finetuning}
Jungo Kasai, Hao Peng, Yizhe Zhang, Dani Yogatama, Gabriel Ilharco, Nikolaos Pappas, Yi~Mao, Weizhu Chen, and Noah~A. Smith.
\newblock Finetuning pretrained transformers into {RNN}s.
\newblock In \emph{Proceedings of the 2021 Conference on Empirical Methods in Natural Language Processing}, pp.\  10630--10643, Online and Punta Cana, Dominican Republic, November 2021. Association for Computational Linguistics.
\newblock \doi{10.18653/v1/2021.emnlp-main.830}.
\newblock URL \url{https://aclanthology.org/2021.emnlp-main.830}.

\bibitem[Katharopoulos et~al.(2020)Katharopoulos, Vyas, Pappas, and Fleuret]{katharopoulos2020transformers}
Angelos Katharopoulos, Apoorv Vyas, Nikolaos Pappas, and Fran{\c{c}}ois Fleuret.
\newblock Transformers are rnns: Fast autoregressive transformers with linear attention.
\newblock In \emph{International conference on machine learning}, pp.\  5156--5165. PMLR, 2020.

\bibitem[Keles et~al.(2023)Keles, Wijewardena, and Hegde]{keles2023computational}
Feyza~Duman Keles, Pruthuvi~Mahesakya Wijewardena, and Chinmay Hegde.
\newblock On the computational complexity of self-attention.
\newblock In \emph{International Conference on Algorithmic Learning Theory}, pp.\  597--619. PMLR, 2023.

\bibitem[Mao(2022)]{mao-2022-fine}
Huanru~Henry Mao.
\newblock Fine-tuning pre-trained transformers into decaying fast weights.
\newblock In \emph{Proceedings of the 2022 Conference on Empirical Methods in Natural Language Processing}, pp.\  10236--10242, Abu Dhabi, United Arab Emirates, December 2022. Association for Computational Linguistics.
\newblock \doi{10.18653/v1/2022.emnlp-main.697}.
\newblock URL \url{https://aclanthology.org/2022.emnlp-main.697}.

\bibitem[Merity et~al.(2017)Merity, Xiong, Bradbury, and Socher]{merity2017pointer}
Stephen Merity, Caiming Xiong, James Bradbury, and Richard Socher.
\newblock Pointer sentinel mixture models.
\newblock In \emph{International Conference on Learning Representations}, 2017.
\newblock URL \url{https://openreview.net/forum?id=Byj72udxe}.

\bibitem[Olsson et~al.(2022)Olsson, Elhage, Nanda, Joseph, DasSarma, Henighan, Mann, Askell, Bai, Chen, et~al.]{olsson2022context}
Catherine Olsson, Nelson Elhage, Neel Nanda, Nicholas Joseph, Nova DasSarma, Tom Henighan, Ben Mann, Amanda Askell, Yuntao Bai, Anna Chen, et~al.
\newblock In-context learning and induction heads.
\newblock \emph{arXiv preprint arXiv:2209.11895}, 2022.

\bibitem[Peng et~al.(2021)Peng, Pappas, Yogatama, Schwartz, Smith, and Kong]{peng2021random}
Hao Peng, Nikolaos Pappas, Dani Yogatama, Roy Schwartz, Noah~A Smith, and Lingpeng Kong.
\newblock Random feature attention.
\newblock \emph{arXiv preprint arXiv:2103.02143}, 2021.

\bibitem[Poli et~al.(2023)Poli, Massaroli, Nguyen, Fu, Dao, Baccus, Bengio, Ermon, and R{\'e}]{poli2023hyena}
Michael Poli, Stefano Massaroli, Eric Nguyen, Daniel~Y Fu, Tri Dao, Stephen Baccus, Yoshua Bengio, Stefano Ermon, and Christopher R{\'e}.
\newblock Hyena hierarchy: Towards larger convolutional language models.
\newblock \emph{arXiv preprint arXiv:2302.10866}, 2023.

\bibitem[Qin et~al.(2022{\natexlab{a}})Qin, Han, Sun, Li, Kong, Barnes, and Zhong]{qin-etal-2022-devil}
Zhen Qin, Xiaodong Han, Weixuan Sun, Dongxu Li, Lingpeng Kong, Nick Barnes, and Yiran Zhong.
\newblock The devil in linear transformer.
\newblock In \emph{Proceedings of the 2022 Conference on Empirical Methods in Natural Language Processing}, pp.\  7025--7041, Abu Dhabi, United Arab Emirates, December 2022{\natexlab{a}}. Association for Computational Linguistics.
\newblock \doi{10.18653/v1/2022.emnlp-main.473}.
\newblock URL \url{https://aclanthology.org/2022.emnlp-main.473}.

\bibitem[Qin et~al.(2022{\natexlab{b}})Qin, Sun, Deng, Li, Wei, Lv, Yan, Kong, and Zhong]{qin2022cosformer}
Zhen Qin, Weixuan Sun, Hui Deng, Dongxu Li, Yunshen Wei, Baohong Lv, Junjie Yan, Lingpeng Kong, and Yiran Zhong.
\newblock cosformer: Rethinking softmax in attention.
\newblock \emph{arXiv preprint arXiv:2202.08791}, 2022{\natexlab{b}}.

\bibitem[Radford et~al.(2019)Radford, Wu, Child, Luan, Amodei, and Sutskever]{Radford2019LanguageMA}
Alec Radford, Jeff Wu, Rewon Child, David Luan, Dario Amodei, and Ilya Sutskever.
\newblock Language models are unsupervised multitask learners.
\newblock 2019.
\newblock URL \url{https://api.semanticscholar.org/CorpusID:160025533}.

\bibitem[Rahimi \& Recht(2007)Rahimi and Recht]{rahimi2007rff}
Ali Rahimi and Benjamin Recht.
\newblock Random features for large-scale kernel machines.
\newblock In J.~Platt, D.~Koller, Y.~Singer, and S.~Roweis (eds.), \emph{Advances in Neural Information Processing Systems}, volume~20. Curran Associates, Inc., 2007.
\newblock URL \url{https://proceedings.neurips.cc/paper_files/paper/2007/file/013a006f03dbc5392effeb8f18fda755-Paper.pdf}.

\bibitem[Schlag et~al.(2021)Schlag, Irie, and Schmidhuber]{schlag2021linear}
Imanol Schlag, Kazuki Irie, and J{\"u}rgen Schmidhuber.
\newblock Linear transformers are secretly fast weight programmers.
\newblock In \emph{International Conference on Machine Learning}, pp.\  9355--9366. PMLR, 2021.

\bibitem[Tay et~al.(2021)Tay, Dehghani, Abnar, Shen, Bahri, Pham, Rao, Yang, Ruder, and Metzler]{tay2021long}
Yi~Tay, Mostafa Dehghani, Samira Abnar, Yikang Shen, Dara Bahri, Philip Pham, Jinfeng Rao, Liu Yang, Sebastian Ruder, and Donald Metzler.
\newblock Long range arena : A benchmark for efficient transformers.
\newblock In \emph{International Conference on Learning Representations}, 2021.
\newblock URL \url{https://openreview.net/forum?id=qVyeW-grC2k}.

\bibitem[Touvron et~al.(2023)Touvron, Martin, Stone, Albert, Almahairi, Babaei, Bashlykov, Batra, Bhargava, Bhosale, et~al.]{touvron2023llama}
Hugo Touvron, Louis Martin, Kevin Stone, Peter Albert, Amjad Almahairi, Yasmine Babaei, Nikolay Bashlykov, Soumya Batra, Prajjwal Bhargava, Shruti Bhosale, et~al.
\newblock Llama 2: Open foundation and fine-tuned chat models.
\newblock \emph{arXiv preprint arXiv:2307.09288}, 2023.

\bibitem[Tsai et~al.(2019)Tsai, Bai, Yamada, Morency, and Salakhutdinov]{tsai-etal-2019-transformer}
Yao-Hung~Hubert Tsai, Shaojie Bai, Makoto Yamada, Louis-Philippe Morency, and Ruslan Salakhutdinov.
\newblock Transformer dissection: An unified understanding for transformer{'}s attention via the lens of kernel.
\newblock In \emph{Proceedings of the 2019 Conference on Empirical Methods in Natural Language Processing and the 9th International Joint Conference on Natural Language Processing (EMNLP-IJCNLP)}, pp.\  4344--4353, Hong Kong, China, November 2019. Association for Computational Linguistics.
\newblock \doi{10.18653/v1/D19-1443}.
\newblock URL \url{https://aclanthology.org/D19-1443}.

\bibitem[Vaswani et~al.(2017)Vaswani, Shazeer, Parmar, Uszkoreit, Jones, Gomez, Kaiser, and Polosukhin]{vaswani2017attention}
Ashish Vaswani, Noam Shazeer, Niki Parmar, Jakob Uszkoreit, Llion Jones, Aidan~N Gomez, {\L}ukasz Kaiser, and Illia Polosukhin.
\newblock Attention is all you need.
\newblock \emph{Advances in neural information processing systems}, 30, 2017.

\bibitem[Warstadt et~al.(2019)Warstadt, Singh, and Bowman]{warstadt2019neural}
Alex Warstadt, Amanpreet Singh, and Samuel~R Bowman.
\newblock Neural network acceptability judgments.
\newblock \emph{Transactions of the Association for Computational Linguistics}, 7:\penalty0 625--641, 2019.

\bibitem[Xiong et~al.(2021)Xiong, Zeng, Chakraborty, Tan, Fung, Li, and Singh]{xiong2021nystromformer}
Yunyang Xiong, Zhanpeng Zeng, Rudrasis Chakraborty, Mingxing Tan, Glenn Fung, Yin Li, and Vikas Singh.
\newblock Nystr{\"o}mformer: A nystr{\"o}m-based algorithm for approximating self-attention.
\newblock In \emph{Proceedings of the AAAI Conference on Artificial Intelligence}, volume~35, pp.\  14138--14148, 2021.

\bibitem[Zheng et~al.(2023)Zheng, Yuan, Wang, and Kong]{zheng2023efficient}
Lin Zheng, Jianbo Yuan, Chong Wang, and Lingpeng Kong.
\newblock Efficient attention via control variates.
\newblock In \emph{The Eleventh International Conference on Learning Representations}, 2023.
\newblock URL \url{https://openreview.net/forum?id=G-uNfHKrj46}.

\end{thebibliography}
\bibliographystyle{iclr2024_conference}
\newpage
\appendix

\section{\name{} implementation details}
\label{app:hh_implement_deets}
\vspace{-0.125cm}
We provide further details on the \name{} feature map and attention weight distillation training.

\subsection{Mechanics for \name{} feature map}
\vspace{-0.125cm}
\label{sec:appendix_mechanics}

To improve \name{} performance in practice, we explored variations along two additional criteria for numerical stability and improved expressivity. 

\textbf{Numerical stability}
In practice, we find that computing $\Phi$ as the softmax applied over the \emph{MLP output dimension} also seems to work but with better stability. In this case, we expand Eq.~\ref{eq:expanded_hh} as
\begin{equation}
    \phi_{\text{mlp}}(\bm{x}) = \Big[\frac{\exp(\bm{w}_1^\top \bm{x})}{\sum_{i=1}^d \exp(\bm{w}_i^\top \bm{x}) }, \ldots, \frac{\exp(\bm{w}_d^\top \bm{x})}{\sum_{i=1}^d \exp(\bm{w}_i^\top \bm{x}) }\Big]
\end{equation}
(also performing better than dividing each element by the max over $\{\exp(\bm{w}_i^\top \bm{x} + \bm{b})\}_{i=1}^d$)

\textbf{Negation mapping}. To better compute dot products as a similarity measure between queries and keys, in practice we also set $\Phi$ as a mapping from $\mathbb{R}^d \mapsto \mathbb{R}^{2d}$, \eg{} via
\begin{equation}
    \phi_{\text{mlp}}(\bm{x}) = \Big[\exp(\bm{w}_1^\top \bm{x} + \bm{b}), \ldots, \exp(\bm{w}_d^\top \bm{x}  + \bm{b}), \exp(-\bm{w}_1^\top \bm{x}  - \bm{b}), \ldots, \exp(-\bm{w}_d^\top \bm{x} - \bm{b}) \Big]
\end{equation}
where the additional negation mapping in $\mathbb{R}^{2d}$ intuitively lets us better factor in negative dimensionalities, which prior linear attention feature maps like $\text{ReLU}$ ignore. While this results in a larger feature dimension, it only scales by a fixed constant $2$ such that the overall time and space complexity for \name{} linear attention is still $\mathcal{O}(nd^2)$. We further find that in practice, this still accomplishes favorable scaling and much faster inference with smaller memory than the Taylor exponential discussed in Sec.~\ref{sec:softmax_approx} (see Fig.~\ref{fig:scaling} for real-world wall-clock time and memory savings).

\subsection{\update{\name{} feature map and model architecture}}
\label{app:arch_fmap_deets}
We apply \name{} feature maps for each head and layer individually in a standard Transformer architecture, where the addition of head-specific MLPs is akin to inserting ``adapters''~\citep{houlsby2019parameter} after every query and key projection. Each MLP is a single linear layer with input and output dimensions equal to the base Transformer's head dimension. Pytorch-like code is given below.\\

\begin{lstlisting}[label={lst:hh_code},language=Python,frame=single,]
import torch
import torch.nn as nn
    
class HedgehogFeatureMap(nn.Module):
    def __init__(self, head_dim: int, activation: str = 'exp'):
        super().__init__()
        # Trainable map
        self.layer = nn.Linear(head_dim, head_dim)
        self.init_weights_()

    def self.init_weights_(self):
        """Initialize trainable map as identity"""
        nn.init.eye_(self.layer.weight)
        nn.init.zeros_(self.layer.bias)
        
    def forward(self, x: torch.Tensor):
        x = self.layer(x)  # shape b, h, l, d
        return torch.cat([torch.exp(x), torch.exp(-x)], dim=-1)
    
\end{lstlisting}

\subsection{\update{\name{} Distillation and Finetuning Implementation Details}}
\label{app:arch_train_deets}
We include additional details for training \name{} layers to obtain linear attention Transformers. These fall under two categories: (1) training-from-scratch, and (2) finetuned / pretrained conversion.

\textbf{1. Training-from-scratch}. When training \name{} Transformers from scratch, we insert a \name{} MLP for each query and key projection of the randomly initialized Transformer (\eg{} for each head of a multi-head attention layer, and for all such layers). We then train the \name{} MLPS jointly with all other model parameters end-to-end with a single objective function, \eg{} cross-entropy loss on next-token prediction when training models for language modeling.

\textbf{2. Finetuned / pretrained conversion}. For both these regimes, we carry out training as a two stage process. Like training-from-scratch, we initially insert \name{} MLPs for query and key projections. Following this, we proceed in two stages:

\begin{enumerate}[leftmargin=*]
    \item \textbf{Attention distillation.}  We first freeze the Transformer's original weights and specifically train the \name{} MLPs, such that the resulting linear attention weights match those produced via softmax attention over the same query and key tensors. For each head, we conceptually follow Listing~\ref{lst:train_fmap} below to compute a soft cross-entropy or KL-divergence between the ``predicted'' linear attention weights and ``ground-truth'' softmax attention weights. 
    We compute these losses for each attention head and layer after one forward pass of the entire model, using data samples from the target task. We find it sufficient to use one optimizer for joint training over all \name{} layers in parallel, using the sum of each individual attention head distillation loss as the final criterion. This makes training simple and comparable to a standard training loop; we further provide code$^\dagger$ in Listing~\ref{lst:hh_arch_for_training} to do so with popular APIs such as HuggingFace Transformers\footnote{\href{https://huggingface.co/docs/transformers/index}{https://huggingface.co/docs/transformers/index}}.

    \item \textbf{Original parameter finetuning.} Following attention distillation, we simply unfreeze all model weights and train with a standard task-specific loss function. We find we can also keep certain layers frozen or train with parameter-efficient finetuning such as low-rank adaptation~\citep{hu2021lora}; we explore this in Sec.~\ref{sec:results_from_pretrained} with Llama-2 models.
\end{enumerate}

\begin{lstlisting}[caption={Hedgehog distillation loss for one attention head}, label={lst:train_fmap},frame=single,language=Python]
# Hedgehog distillation loss for one attention head

def softmax_attn(q: torch.Tensor, k: torch.Tensor):
    """Get softmax attention weights -> Assume q, k are both shape (b, h, l, d)"""
    scale = q.shape[-1] ** 0.5
    qk = torch.einsum('bhmd,bhnd->bhmn', q, k) / scale
    return torch.softmax(qk, dim=-1)

def quadratic_linear_attn(q: torch.Tensor, k: torch.Tensor):
    """
    Get linear attention weights 
    -> Assume q, k are both shape (b, h, l, d), and feature maps already applied
    """
    qk = torch.einsum('bhmd,bhnd->bhmn', q, k)
    return qk / qk.sum(dim=-1, keepdim=True)

def compute_hedgehog_loss(q: torch.Tensor, 
                          k: torch.Tensor,
                          hh_mlp_q: HedgehogFeatureMap, 
                          hh_mlp_k: HedgehogFeatureMap):
    """
    Compute the attention distillation loss
    -> Assume `soft_label_cross_entropy` is implemented 
       (alternatively use KL divergence)
    -> Assume q and k are the queries and keys of a 
       pretrained Transformer, 
       e.g., via q = self.q_proj(hidden_states)
    """
    true_attn = softmax_attn(q, k)
    pred_attn = quadratic_linear_attn(hh_mlp_q(q), hh_mlp_k(k))
    return soft_label_cross_entropy(pred_attn, true_attn)
        \end{lstlisting}

\begin{lstlisting}[caption={Hedgehog Attention class for easy attention distillation.}, label={lst:hh_arch_for_training},frame=single,language=Python]
# Hedgehog Attention class for easy attention distillation

class HedgehogAttention(nn.Module):
    """
    Sample code for HedgehogAttention, following HuggingFace API
    """
    def __init__(self, base_attn, training = True):
        self.base_attn = base_attn  # e.g., LlamaAttention

        # Trainable feature maps
        self.mlp_q = HedgehogFeatureMap(base_attn.head_dim)
        self.mlp_k = HedgehogFeatureMap(base_attn.head_dim)

        # Freeze original attention parameters
        for p in self.base_attn.parameters():
            p.requires_grad = False
            
        self.q_proj = self.base_attn.q_proj
        self.k_proj = self.base_attn.k_proj

        # Whether we train attentions or not
        self.training = training   

    def forward(self, 
                hidden_states: torch.Tensor,
                output_attentions: bool = True,
                **base_kwargs: any):
                
        if self.training:  
            # Compute ground-truth attention weights
            outputs, true_attns = self.base_attn(
                hidden_states=hidden_states,
                output_attentions=True,
                **base_kwargs)

            # Compute Hedghog feature maps
            q = self.mlp_q(self.q_proj(hidden_states))
            k = self.mlp_k(self.k_proj(hidden_states))

            pred_attns = quadratic_linear_attn(q, k)

            if output_attentions:  # Hook for attentions
                return outputs, (pred_attns, true_attns)
      
       # ... End relevant
       
\end{lstlisting}

$^\dagger$In practice, to train all \name{} layers easily in a joint end-to-end fashion, we make use of popular pretrained Transformer APIs such as those in the HuggingFace transformers library. We implement a \name{} equivalent of the base Transformers' attention class, which (1) abstracts away the Transformer-specific attention computation and (2) lets us hook attention weights calculated at each layer to the model's final outputs, \eg{} via \texttt{output\_attentions = True} keyword args.
We can subsequently substitute each attention layer with the ``HedgehogAttention'' equivalent, and train via a simple loop over the data. We present Pytorch-like code in Listing~\ref{lst:hh_training_loop}.

\newpage

\begin{lstlisting}[caption={End-to-end joint attention distillation.},frame=single, label={lst:hh_training_loop},language=Python]
# End-to-end joint attention distillation

from transformers import AutoModel

# Load base model 
base_model = AutoModel.from_pretrained(...)

# Freeze original parameters
for p in base_model: p.requires_grad = False

# Convert attentions for all layers
for layer in base_model:
    base_model.attn = HedgehogAttention(base_model.attn)

# Define single optimizer for training all feature maps
optim = optimizer(base_model.parameters())

# Train Hedgehog feature maps
for data in dataloader:

    # Compute outputs and hook to attentions
    outputs = base_model(**data, output_attentions=True)
    outputs = outputs.get('attentions')
    
    total_loss = 0
    for attns in enumerate(outputs):  # attentions for each layer
        pred_attn, true_attn = attns
        total_loss += soft_label_cross_etnropy(pred_attn, true_attn)
        
    loss.backward()  # Jointly optimize all feature maps
    optim.step()

\end{lstlisting}


\section{Deferred experimental details}

\subsection{Associative recall analysis (Section~\ref{sec:analysis_experiments})}
\label{app:ar_analysis}

In Sec.~\ref{sec:analysis_experiments}, we compare various Transformers' abilities to solve Associative Recall (AR)~\citep{ba2016using}, a next-token prediction task previously studied as a proxy for language modeling capability~\citep{olsson2022context}. AR tests how well a model can recall specific content in an input sequence, structured as a list of key-value pairs which ends in a key Table~\ref{table:associative_recall}. 

\begin{table}[ht]
\begin{center}
\begin{tabular}{@{}cccc@{}}
\toprule
\multirow{2}{*}{Input Sequence} & Next & Vocab & Seq. \\ 
 & Token & Size & Length\\
 \midrule 
\textcolor{blue}{c} \underline{9} k 8 j 3 ... f 1 \textcolor{blue}{c}    & 9          & 40         & 128             \\ \bottomrule
\end{tabular}
\end{center}
\caption{Associative recall task. Example from~\cite{ba2016using}.} 
\label{table:associative_recall}
\end{table}

\paragraph{Dataset details.}
To understand the effects of more uniform attention weightings, we evaluate with 40 possible tokens and 128 token-long-sequences, such that models must recall pairings that only occur three times on average in-context. We generate 10,000 training samples following the patterns described in Table~\ref{table:associative_recall}, and evaluate on 2000 newly-generated test samples (again using the same associative recall structure, but with different token associations).

\paragraph{Architecture details.}
For all experiements, we use a four layer Transformer with four heads-per-layer, head dimension = 64, and rotary embeddings. This is similar to modern model families such as Pythia~\citep{biderman2023pythia} and LLaMA / Llama-2~\citep{touvron2023llama}. We keep all parts consistent except for the multi-head attention, comparing popular linear attentions (c.f., Fig.~\ref{fig:entropy_viz}).

\paragraph{Training details.}
For fair comparison to evaluate just the feature map / modeling architecture, we train all models by sweeping learning rate $\in$ \{1e-2, 1e-4\}, weight decay $\in$ \{0, 5e-4\}, and batch size $\in$ \{8, 32\} with AdamW optimizer. We train up to 100 epochs with early stopping (explicitly stopping training if validation loss stops decreasing after 10 epochs).

\subsection{BERT-base finetuned on CoLA conversion (Section~\ref{sec:analysis_experiments})}
\label{app:bert_ft_analysis}

\paragraph{Training details.} For our finetuned-conversion analysis, we replace the attentions of a finetuned BERT-base-uncased model available on the HuggingFace model hub\footnote{\href{https://huggingface.co/JeremiahZ/bert-base-uncased-cola}{https://huggingface.co/JeremiahZ/bert-base-uncased-cola}}. We train with batch size 8, learning rate 1e-5, zero weight decay, AdamW optimizer, and up to 10 epochs with early stopping.

\subsection{\name{} training from scratch (Section~\ref{sec:results_from_scratch})}
\label{app:train_scratch_deets}

\paragraph{LRA training and model details.} On LRA, for fair comparison we implement \name{} in the existing PyTorch implementation provided by~\cite{xiong2021nystromformer}, deferring to the same model configurations and hyperparameters used in the original repository~\citep{tay2021long}.

\paragraph{WikiText-103 training and model details.} 
For WikiText-103, we train a 125M parameter GPT-2 style Transformer with learning rate 6e-4, weight decay 0.01, and AdamW optimizer. For close comparison, we follow the architectural details of GPT-2 125M, and use a 12 layer decoder-only network with 12 heads, head dimension = 64, hidden dimension 768, and MLP dimension 3072. 

\subsection{\name{} finetuned conversion (Section~\ref{sec:results_from_finetuned})}
\label{app:train_finetuned_deets}

\paragraph{Recovering finetuned BERT performance on GLUE tasks.} For finetuned conversion, we first conduct \name{} attention distillation 
by training attention layers up to five epochs with early stopping based on validation loss. We train with learning rate 1e-2, weight decay 0, AdamW optimizer. We follow the same procedure for the Transformer-to-RNN (T2R)~\citep{kasai-etal-2021-finetuning} ablation. For regulard T2R and subsequently post attention distillation, we train each BERT model with batch size 8, learning rate 1e-5, weight decay 0, AdamW optimizer, and cosine scheduler for up to five epochs on the individual classification (all except STS-B) or regression tasks (STS-B) on the GLUE benchmark. 
For all tasks, we use the corresponding available finetuned BERT-base-uncased checkpoints hosted at the HuggingFace model hub\footnote{\href{https://huggingface.co/JeremiahZ/}{https://huggingface.co/JeremiahZ/}}, and thank the original uploader for their contributions.

\paragraph{Recovering finetuned Vision Transformer performance on ImageNet-1K.} \update{To demonstrate finetuned-conversion for the image domain, we use the \texttt{vit-base-patch16-224} checkpoint provided by Google on HuggingFace\footnote{\href{https://huggingface.co/google/vit-base-patch16-224}{https://huggingface.co/google/vit-base-patch16-224}}, which is trained on ImageNet-21k before being finetuned on ImageNet-1K at resolution of 224 x 224 pixels~\citep{dosovitskiy2020image}. For distillation, we freeze the original ViT weights, and train linear attention MLPS with batch size 32, learning rate 0.01, zero weight decay, and AdamW optimizer, and train for two epochs. We then train all parameters with learning rate 1e-3, zero weight decay and AdamW optimizer up to 10 epochs with early stopping.}

\subsection{\name{} pretrained conversion (Section~\ref{sec:results_from_pretrained})}
\label{app:train_pretrained_deets}

\paragraph{Linear GPT-2 125M conversion for WikiText-103 language modeling.} We use the available GPT-2 125M pretrained checkpoint available on HuggingFace\footnote{\href{https://huggingface.co/gpt2}{https://huggingface.co/gpt2}} from~\cite{Radford2019LanguageMA}. For \name{}, we first do attention distillation and train \name{} MLPs for two epochs over the WikiText-103 data, using batch size 8, learning rate 0.01, zero weight decay, AdamW optimizer, and 1024-tokens per input. For T2R-GPT-2 and the subsequent \name{}-GPT-2 model, we finetune all model parameters with learning rate 6e-4, weight decay 0.01, and AdamW optimizer and 1024 tokens-per-input.

\paragraph{Linear Llama-2 7B conversion for SAMSum corpus summarization.} \update{ We use the base Llama-2 7B model available via Meta and HuggingFace (\texttt{llama-2-7b-hf}) from~\cite{touvron2023llama}. For all experiments, we use non-quantized model weights in bfloat16, and conduct all training runs and evaluations on a single A6000 GPU.}

\update{For dataset preparation, we first convert individual document and summarization pairs into single next-token prediction samples, using the template in Listing~\ref{lst:samsum_llama_prompt}. For both distillation and subsequent finetuning, we then chunk these samples into concatenated inputs 1024 tokens long.} 

\update{For attention distillation, we freeze all original Llama-2 weights, and train \name{} MLPs for every head and layer (0.495\% of the original model size). We then train for two epochs with learning rate 0.01, zero weight decay, AdamW optimizer, and batch size 8 with gradient accumulation. }

\update{For finetuning and comparison to T2R and standard attention, we apply LoRA to query, key, value, and output projections of each layer. We use alpha parameter 16 and rank 8. We train with learning rate 1e-4, zero weight decay, AdamW optimizer, and batch size 8 with gradient accumulation.}

\update{For generation, we compute ROUGE metrics (R1, R2, RL; for overlap of unigrams, bigrams, and longest common subsequence) over model outputs. We generate sequences up to 100 tokens long, and evaluate based on outputs up til the first \texttt{</s>} Llama stop token.}

\vspace{0.25cm}
\begin{lstlisting}[caption={Llama-2 prompt template for SAMSum corpus summarization},frame=single, label={lst:samsum_llama_prompt},language=Python]
# Llama-2 prompt template for SAMSum

Summarize this dialog:
{input}
---
Summary:
{output}{eos_token}
\end{lstlisting}

\section{Additional results}
\label{app:add_results}

\subsection{Extended comparison to attention models on LRA}
\label{app:LRA_full}

In Table~\ref{table:LRA_full}, we compare \name{}'s performance on LRA against a fuller set of Transformer and subquadratic Transformer based alternatives sourced either from the official benchmark leaderboard~\citep{tay2021long} or recent subquadratic attention works (where we display the most competitive alternatives in Table~\ref{table:LRA}). We find \name{} on average obtains best accuracy. \update{Although recently non-Transformer models such as deep state-space models have shown impressive results outperforming Transformers on LRA~\citep{gu2021efficiently}, as our work focuses on how to improve and recover the expressivity of standard softmax Transformers, we focus the comparison against other attention-based methods. We defer to~\cite{gu2021efficiently} and related works for their LRA  results.}

{\small
\begin{table}[ht]
\label{table:LRA_app}
\begin{center}
\begin{tabular}{@{}lcccccc@{}}
\toprule
Model           & ListOps        & Text           & Retrieval      & Image          & Pathfinder     & Average            \\ \midrule
Transformer     & 36.37          & 64.27          & 57.46          & 42.44          & 71.40          & 54.39          \\ \midrule
Local Att       & 15.82          & 52.98          & 53.39          & 41.46          & 66.63          & 46.06          \\
Linear Trans.   & 16.13          & 65.90          & 53.09          & \underline{42.34}    & \underline{75.30}    & 50.55          \\
Reformer        & 37.27          & 56.10          & 53.40          & 38.07          & 68.50          & 50.67          \\
Sparse Trans.   & 17.07          & 63.58          & 59.59          & 44.24          & 71.71          & 51.24          \\
Sinkhorn Trans. & 33.67          & 61.20          & 53.83          & 41.23          & 67.45          & 51.29          \\
Linformer       & 35.70          & 53.94          & 52.27          & 38.56          & 76.34          & 51.36          \\
Performer       & 18.01          & 65.40          & 53.82          & \textbf{42.77} & \textbf{77.05} & 51.41          \\
Synthesizer     & 36.99          & 61.68          & 54.67          & 41.61          & 69.45          & 52.88          \\
Longformer      & 35.63          & 62.85          & 56.89          & 42.22          & 69.71          & 53.46          \\
BigBird         & 36.05          & 64.02          & 59.29          & 40.83          & 74.87          & 55.01          \\
Nyströmformer$^\dagger$   & 37.15          & 65.52          & 79.56          & 41.58          & 70.94          & 58.95          \\
cosFormer$^\dagger$       & \underline{37.90}    & 63.41          & 61.36          & 43.17          & 70.33          & 55.23          \\
Skyformer$^\dagger$       & \textbf{39.25} & 64.70  & \underline{82.06}    & 40.77          & 70.73          & \underline{59.50}    \\  \midrule
Hedgehog        & 37.15          & \underline{64.60}    & \textbf{82.24} & 40.15          & 74.16          & \textbf{59.66} \\ \bottomrule
\end{tabular}
\end{center}
\caption{Training-from-scratch on LRA. \name{} achieves best average performance across Transformers and subquadratic variants. $^\dagger$ indicates method results reported from original works. All other reported from the official LRA benchmark~\citep{tay2021long}. \textbf{Best}, \underline{2nd-best} acc (\%).} 
\label{table:LRA_full}
\end{table}
}

\subsection{\update{\name{} feature map generalization to new data}}
\label{app:add_fmap_gen}
We extend our analysis into how \name{}'s feature maps learned with one dataset generalize to attentions computed on a new dataset (\emph{c.f.} Table~\ref{table:generalize_kl} and Fig.~\ref{fig:generalize_data_viz} in Sec.~\ref{sec:hedgehog_benchmarking}). As in the prior section, we find \name{} learned feature maps frequently generalize to new datasets. Despite training to match the softmax attentions on one model and dataset, we first find \name{} feature maps can produce attention weights that closely resemble softmax attention for the same model on another dataset (App.~\ref{app:add_gen_data_viz}). We next quantify this fidelity via KL divergence w.r.t. the softmax attentions (App.~\ref{app:add_gen_kl}). We find that \name{} learned feature maps almost always still generalize better than prior linear attention feature maps. We finally show that this attention matching generalization transfers to actual pretrained-conversion performance (App.~\ref{app:add_gen_downstream_eval}). We replace BERT-base softmax attentions with \name{} attentions trained on one task, and find finetuning with these converted models on \emph{another} GLUE task still leads to improvements over prior linear attentions. 

\paragraph{Setup.} For all experiments, we begin by training \name{} attentions on ``in-distribution'' softmax attention data. We use the pretrained BERT-base-uncased model~\citep{devlin2018bert} as the Transformer we wish to convert, and distill two sets of \name{} attentions over (1) the GLUE CoLA task or (2) 512-token chunks of WikiText-103 corpus. Thus, queries and keys computed with the BERT-base-uncased model over CoLA validation samples are ``in-distribution'' for the first set, and we are interested in seeing how attention weight fidelity or downstream performance recovery are affected when subsequently finetuning on non-CoLA GLUE data. We compare with various prior ablations and alternative feature maps, such as the Transformer-to-RNN feature map~\citep{kasai-etal-2021-finetuning} after attention distillation, \name{} without attention distillation, and prior representative linear attentions such as Performer~\citep{choromanski2020rethinking} and cosFormer~\citep{qin2022cosformer}.



\subsubsection{\update{Qualitative evidence of \name{} data generalization}}\label{app:add_gen_data_viz}

\begin{figure}[t!]
    \centering
    \includegraphics[width=1\linewidth]{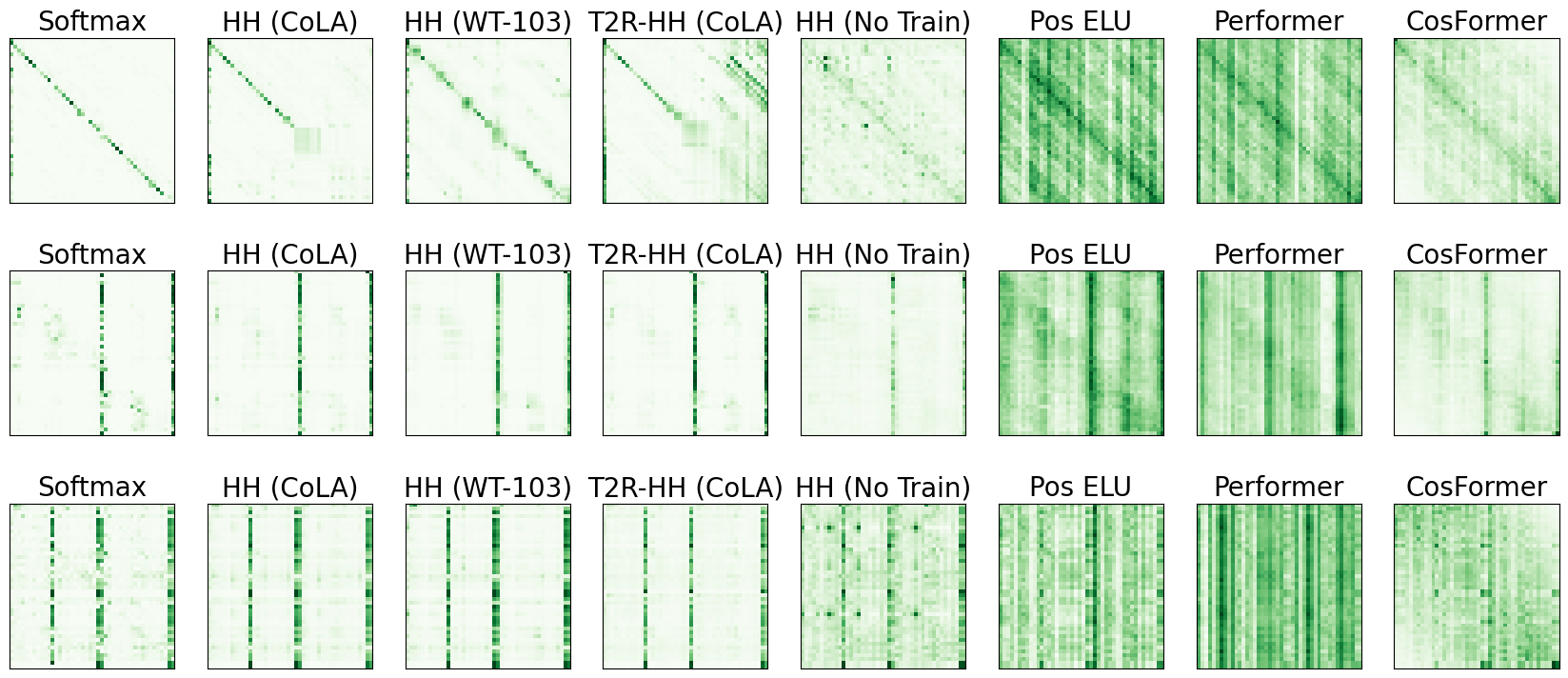}
    \caption{\textbf{Qualitative generalization to MRPC}. Attention weights for BERT-base-uncased queries and keys computed on MRPC samples. We compare attentions from the 3rd head in the 1st, 6th and 12th layers (top, middle, bottom). \name{} feature maps trained on CoLA or WikiText-103 often still produce attention weights similar to those of softmax attention on new data.}
    \label{fig:cola_mrpc}
\end{figure}

\begin{figure}[t!]
    \centering
    \includegraphics[width=1\linewidth]{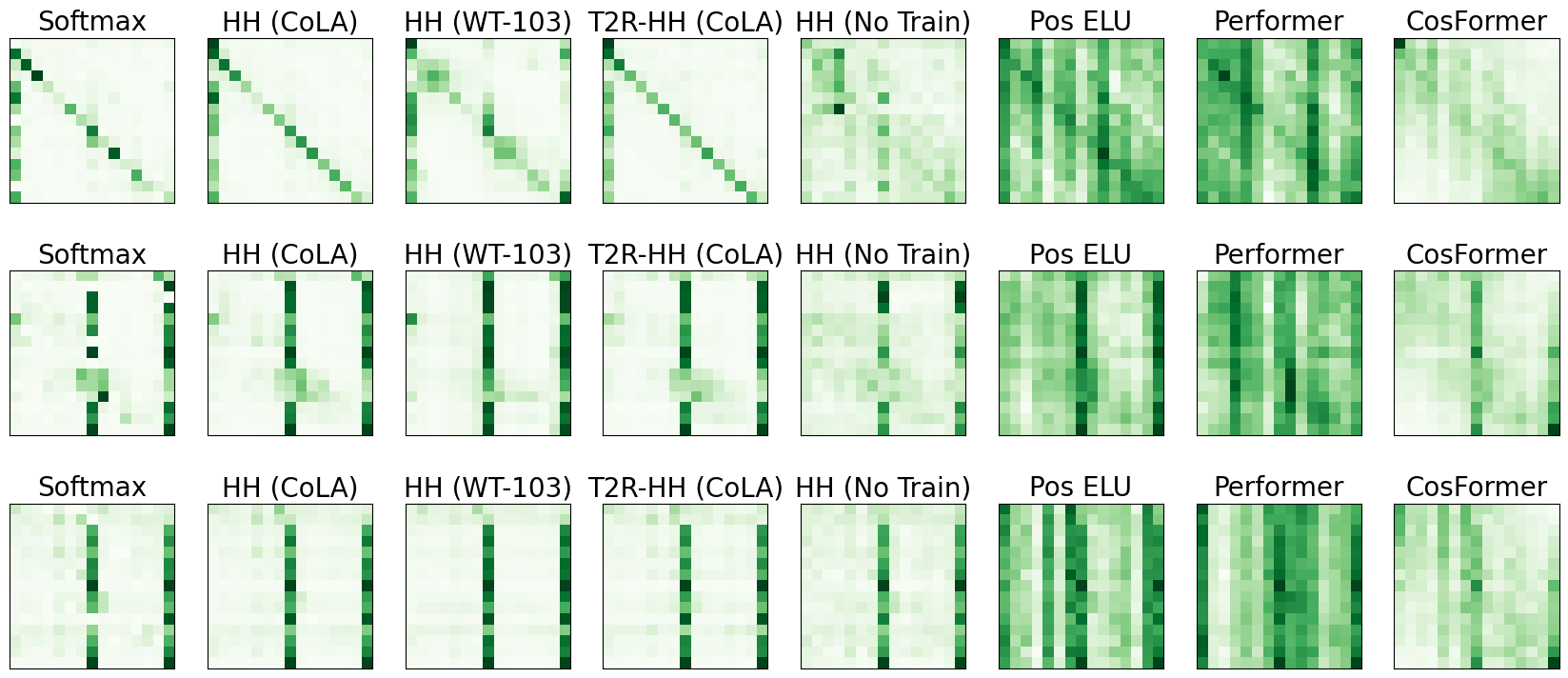}
    \caption{\textbf{Qualitative generalization to QNLI}. Attention weights for BERT-base-uncased queries and keys computed on QNLI samples. We compare attentions from the 3rd head in the 1st, 6th and 12th layers (top, middle, bottom). \name{} feature maps trained on CoLA or WikiText-103 often still produce attention weights similar to those of softmax attention on new data.}
    \label{fig:cola_qnli}
\end{figure}

In Fig.~\ref{fig:cola_mrpc} and Fig.~\ref{fig:cola_qnli}, we visualize attention weights computed via various methods on heads in the 1st, 6th, and 12th layers of the BERT-base uncased model. We find \name{} can learn feature maps that lead to matching softmax attention weights, even when computed on new data samples. Interestingly, the \name{} feature maps result in significantly more similar attention weights versus alternative feature maps (quantified in the next section).

In addition, our comparisons to \name{} ablations suggest that the proposed \name{} feature map \emph{and} distillation procedure are important for best generalization. Removing either the \name{} feature map form (via doing attention distillation using the prior Transformer-to-RNN feature map (T2R-HH) or not training feature maps (HH (No Train)) leads to lower fidelity, where attention distillation seems critical for retaining weights reasonably similar to softmax attention.

\subsubsection{\update{Quantitative analysis of \name{} data generalization}}
\label{app:add_gen_kl}
To quantify the above observations, we compute the KL divergence between \name{} attention weights computed on various GLUE tasks and the ``ground-truth'' softmax attention weights, using the pretrained BERT-base-uncased model. We report the KL divergence in Table~\ref{table:kl_gen_data}. Similar to the above visualizations, we find that \name{} feature maps do seem to produce better matching attention weights to softmax attention via significantly smaller KL divergences.

\begin{table}[]
\resizebox{\linewidth}{!}{
\begin{tabular}{@{}lcccccccc@{}}
\toprule
Method             & CoLA           & MNLI           & MRPC           & QNLI           & QQP            & RTE            & SST-2          & STS-B          \\ \midrule
Hedgehog (CoLA)     & \textbf{0.173} & \textbf{0.340} & {\ul 0.652}    & {\ul 0.673}    & \textbf{0.374} & {\ul 0.932}    & \textbf{0.211} & \textbf{0.275} \\
Hedgehog (WT-103)   & 0.357          & {\ul 0.381}    & \textbf{0.484} & \textbf{0.446} & {\ul 0.432}    & \textbf{0.428} & {\ul 0.347}    & {\ul 0.360}    \\
T2R-HH (CoLA) & {\ul 0.191}    & 0.603          & 1.124          & 1.141          & 0.658          & 1.357          & 0.254          & 0.444          \\
Hedgehog Untrained  & 0.687          & 0.845          & 1.264          & 1.146          & 0.890          & 1.493          & 0.859          & 0.743          \\ \midrule
1 + ELU             & 1.231          & 1.500          & 2.150          & 1.947          & 1.509          & 2.491          & 1.505          & 1.285          \\
Performer           & 1.293          & 1.592          & 2.239          & 2.059          & 1.588          & 2.594          & 1.558          & 1.352          \\
CosFormer           & 1.191          & 1.341          & 1.987          & 1.840          & 1.330          & 2.398          & 1.443          & 1.142          \\ \bottomrule
\end{tabular}
}
\caption{\textbf{KL divergence of attention weights generalizing to new data}. \name{} attentions trained on either CoLA (CoLA) or WikiText-103 (WT-103) data, still best match softmax attention weights computed on different GLUE tasks, despite being trained with task-specific data (measured via KL divergence; lower is better).} 
\label{table:kl_gen_data}
\end{table}

\subsubsection{\update{\name{} data generalization via GLUE task transfer}}
\label{app:add_gen_downstream_eval}
We finally evaluate the \name{} attention generalization by finetuning the pretrained BERT models with trained \name{} on new GLUE tasks. We follow the same procedure described in Appendix~\ref{app:train_finetuned_deets}. In Table~\ref{table:task_gen_task}, we find that the above attention weight observations on \name{} generalization also correspond with downstream task performance. \name{}-BERT models achieve best or second-best performance, despite using attention feature maps trained on different data. 
We leave further generalization studies, such as how \name{} attentions trained on one model generalize to \emph{an entirely different} model for future work.

\begin{table}[t!]
\resizebox{\linewidth}{!}{
\begin{tabular}{@{}lccccccc@{}}
\toprule
Method            & CoLA          & MRPC          & QNLI          & QQP           & RTE           & SST-2         & STS-B         \\ \midrule
Hedgehog (CoLA)   & \textbf{58.4} & \textbf{89.4} & {\ul 87.7}    & {\ul 89.8}    & {\ul 62.1}    & \textbf{91.9} & {\ul 85.3}    \\
Hedgehog (WT-103) & 47.2          & \textbf{89.4} & \textbf{89.2} & \textbf{90.4} & \textbf{62.5} & {\ul 91.4}    & \textbf{86.7} \\
HH (No Train)     & {\ul 50.3}    & {\ul 83.3}    & 85.9          & 86.5          & 55.6          & 89.5          & 79.3          \\ \midrule
1 + ELU           & 26.8          & 81.9          & 78.5          & 89.1          & 55.6          & 85.9          & 41.8          \\
Performer         & 24.7          & 81.4          & 75.8          & 86.5          & 55.6          & 85.1          & 39.8          \\
CosFormer         & 41.1          & 82            & 82.6          & 89.3          & 54.9          & 88.4          & 76.6          \\ \bottomrule
\end{tabular}
}
\caption{\textbf{Attention generalization on downstream tasks}. BERT models with \name{} attentions trained on either CoLA (CoLA) or WikiText-103 (WT-103) achieve best GLUE performance despite being finetuned on different GLUE tasks. This corresponds with prior observations in generalization via improved attention weight fidelity.} 
\label{table:task_gen_task}
\end{table}

\newpage

\subsection{\update{Llama-2 SAMSum Generations}}
\label{app:llama_gen}
We include sample generations from the SAMSum corpus summarization task~\citep{gliwa-etal-2019-samsum}, used to evaluate \name{} conversion of LLama-2 7B models in combination with low-rank adaptation (LoRA). Via the generation quality, we find that in contrast to prior conversion methods such as Transformer-to-RNN (T2R)~\citep{kasai-etal-2021-finetuning}, \name{} makes pretrained-conversion with parameter-efficient finetuning feasible on larger models. 

We report generations for four test samples of the SAMSum test set (first 3, and a longer 6th), comparing standard attention Llama-2 models, linear attention Llama-2 models achieved via \name{} attention distillation, and linear attention Llama-2 models following T2R, all with LoRA finetuning on a single A6000 GPU (unquantized bfloat16 weights; LoRA on all query, key, value, and output projections).  

For future work, we think further showing how \name{} and pretrained-conversion can help apply larger models to tasks requiring longer contexts or more efficient inference, are promising avenues of exploration. Several considerations include additional explorations to improve generation quality, parameter training efficiency (\eg{} combining \name{} with quantization~\citep{dettmers2023qlora}), and further utilizing \name{}'s linear attention efficiency via transfer to longer contexts (\eg{} combining \name{} with positional interpolation~\citep{chen2023extending}).\\

\begin{lstlisting}[caption={SAMSum Llama-2 generations after LoRA finetuning.}, label={lst:samsum_gen1}, frame=single, numbers=none]
<@\textcolor{gray}{\# (Input):}@>
<s> Summarize this dialog:
Hannah: Hey, do you have Betty's number?
Amanda: Lemme check
Hannah: <file_gif>
Amanda: Sorry, can't find it.
Amanda: Ask Larry
Amanda: He called her last time we were at the park together
Hannah: I don't know him well
Hannah: <file_gif>
Amanda: Don't be shy, he's very nice
Hannah: If you say so..
Hannah: I'd rather you texted him
Amanda: Just text him ��
Hannah: Urgh.. Alright
Hannah: Bye
Amanda: Bye bye
---
Summary:

<@\textcolor{red}{\# (Llama-2):}@>
Hannah is looking for Betty's number. Amanda can't find it. Amanda 
suggests Hannah to ask Larry.</s>

<@\textcolor{blue}{\# (Hedgehog Llama-2):}@>
Hannah is looking for Betty's number. Amanda will text her.</s>

<@\textcolor{gray}{\# (T2R Llama-2):}@>
along recently acquired the biggest, I'tch.ЉSCHPW182seud is the 
biggest, I't a.ЉSCHPentua.ЉSCHPentle't a.ЉSCHPentua.Љ?
along ch.Љ?
along ch.Љ?
along ch.Љ?
along ch.Љ?
along ch.Љ?
along ch.Љ?
\end{lstlisting}

\newpage
\begin{lstlisting}[caption={SAMSum Llama-2 generations after LoRA finetuning.}, label={lst:samsum_gen2}, frame=single, numbers=none]
<@\textcolor{gray}{\# (Input):}@>
<s> Summarize this dialog:
Eric: MACHINE!
Rob: That's so gr8!
Eric: I know! And shows how Americans see Russian ;)
Rob: And it's really funny!
Eric: I know! I especially like the train part!
Rob: Hahaha! No one talks to the machine like that!
Eric: Is this his only stand-up?
Rob: Idk. I'll check.
Eric: Sure.
Rob: Turns out no! There are some of his stand-ups on youtube.
Eric: Gr8! I'll watch them now!
Rob: Me too!
Eric: MACHINE!
Rob: MACHINE!
Eric: TTYL?
Rob: Sure :)
---
Summary:

<@\textcolor{red}{\# (Llama-2):}@>
Eric and Rob are watching a stand-up by a Russian comedian.</s>

<@\textcolor{blue}{\# (Hedgehog Llama-2):}@>
Eric and Rob are watching a stand-up comedy by a Russian comedian.</s>

<@\textcolor{gray}{\# (T2R Llama-2):}@>
along recently acquired the biggest, 
I'tch.ЉSCHPentua.ЉSCHPentua.ЉSCHPentua.ЉSCHPentua.ЉSCHP
along ch.Љ?
along ch.Љ?
along ch.ЉSCHP
along ch.Љ?
along ch.Љ?
along ch.Љ?
along for:
along for:
al
\end{lstlisting}

\newpage
\begin{lstlisting}[caption={SAMSum Llama-2 generations after LoRA finetuning.}, label={lst:samsum_gen3}, frame=single, numbers=none]
<@\textcolor{gray}{\# (Input):}@>
<s> Summarize this dialog:
Lenny: Babe, can you help me with something?
Bob: Sure, what's up?
Lenny: Which one should I pick?
Bob: Send me photos
Lenny:  <file_photo>
Lenny:  <file_photo>
Lenny:  <file_photo>
Bob: I like the first ones best
Lenny: But I already have purple trousers. Does it make sense to have 
two pairs?
Bob: I have four black pairs :D :D
Lenny: yeah, but shouldn't I pick a different color?
Bob: what matters is what you'll give you the most outfit options
Lenny: So I guess I'll buy the first or the third pair then
Bob: Pick the best quality then
Lenny: ur right, thx
Bob: no prob :)
---
Summary:

<@\textcolor{red}{\# (Llama-2):}@>
Lenny is looking for new trousers. Bob suggests the first or the 
third pair.</s>

<@\textcolor{blue}{\# (Hedgehog Llama-2):}@>
Lenny wants to buy a new pair of trousers. Bob suggests the first two 
options.</s>

<@\textcolor{gray}{\# (T2R Llama-2):}@>
along, I't a.ЉSCHPentua.ЉSCHP
along ch.ЉSCHPentua.Љ?
along ch.ЉSCHP
along ch.Љ?
along ch.Љ?
along ch.Љ?
along for:
along ch.Љ?
along ch.Љ?
along for:
along ch.Љ?
along for:

\end{lstlisting}

\newpage
\begin{lstlisting}[caption={SAMSum Llama-2 generations after LoRA finetuning.}, label={lst:samsum_gen4}, frame=single, numbers=none]
<@\textcolor{gray}{\# (Input):}@>
<s> Summarize this dialog:
Benjamin: Hey guys, what are we doing with the keys today?
Hilary: I've got them. Whoever wants them can meet me at lunchtime or 
after
Elliot: I'm ok. We're meeting for the drinks in the evening anyway 
and I guess we'll be going back to the apartment together?
Hilary: Yeah, I guess so
Daniel: I'm with Hilary atm and won't let go of her for the rest of 
the day, so any option you guys choose is good for me
Benjamin: Hmm I might actually pass by at lunchtime, take the keys 
and go take a nap. I'm sooo tired after yesterday
Hilary: Sounds good. We'll be having lunch with some French people 
(the ones who work on the history of food in colonial Mexico - I 
already see you yawning your head off)
Benjamin: YAAAAWN �� Where and where are you meeting?
Hilary: So I'm meeting them at the entrance to the conference hall at 
2 pm and then we'll head to this place called La Cantina. Italian 
cuisine, which is quite funny, but that's what they've chosen
Benjamin: Interesting �� To be honest, Hilary, I almost feel like 
changing my mind. Wanting to take this nap might end up costing me to 
dear
Hilary: Oh come on ��
Benjamin: All these terrible obstacles on mu way to bed might just 
prove to much to take
Hilary: We'll try to avoid talking about their subject of research. 
Oh wait, no, I'm actually meeting them because I wanted to chat about 
their research lol
Elliot: ��
Hilary: Do join us, we're going to have fun. And then you'll take the 
keys and take this most deserved of naps
Elliot: Sounds like a plan ��
Hilary: ��
Elliot: See you at 2 then xx

<@\textcolor{red}{\# (Llama-2):}@>
Hilary has the keys. Benjamin is tired after yesterday. Hilary is 
meeting some French people at 2 pm. Benjamin will join them.</s>

<@\textcolor{blue}{\# (Hedgehog Llama-2):}@>
Hilary and Elliot are going to meet at the entrance of the conference 
hall at 2 pm. Hilary and Benjamin will meet there. Hilary and 
Benjamin will take the keys to the apartment. Hilary and Benjamin 
will meet Elliot at 2 pm. Hilary and Benjamin will take a nap.</s>

<@\textcolor{gray}{\# (T2R Llama-2):}@>
Most is the biggest, I's:
Most is the biggest, I's:
Most is the biggest, I's:
Most is the biggest, I's:
Most is the biggest, I's:
Most is the biggest, I's:
M
\end{lstlisting}

\subsection{\update{Additional attention weight visualizations}}

We finally include additional visualizations of the attention weights computed via softmax attention in comparison to \name{} and alternate linear attention feature maps. We visualize attentions computed on GLUE tasks (Sec.~\ref{sec:results_from_pretrained}) from the 1st, 6th, and 12th (first, middle, last) layers of BERT models in top, middle, and bottom rows respectively, and for the 1st, 6th, and 12th heads.

\newpage

\begin{figure}
    \centering
    \includegraphics[width=0.95\linewidth]{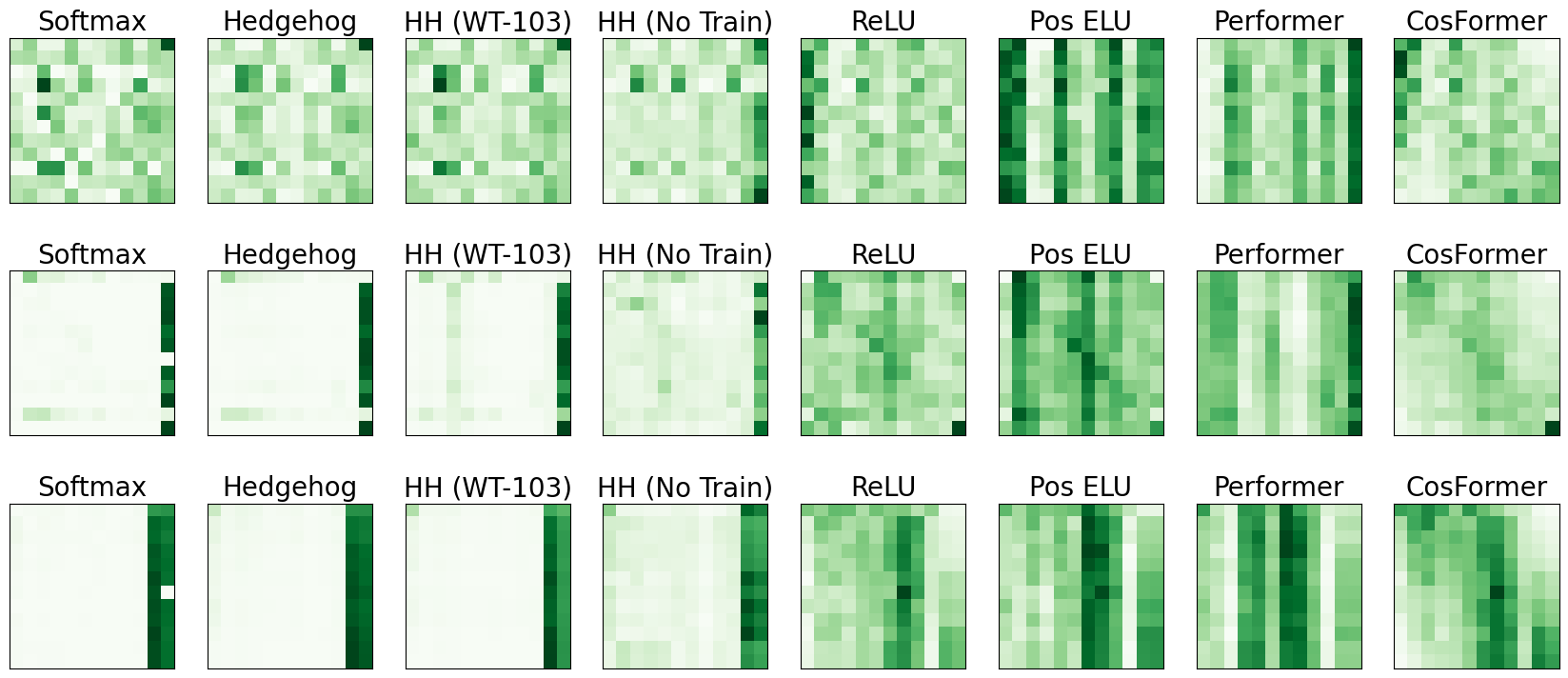}
    \caption{BERT attention visualizations for CoLA. Head 1; 1st, 6th, and 12th layers.}
    \label{fig:enter-label1}
    \vspace{-0.35cm}
\end{figure}

\begin{figure}
    \centering
    \includegraphics[width=0.95\linewidth]{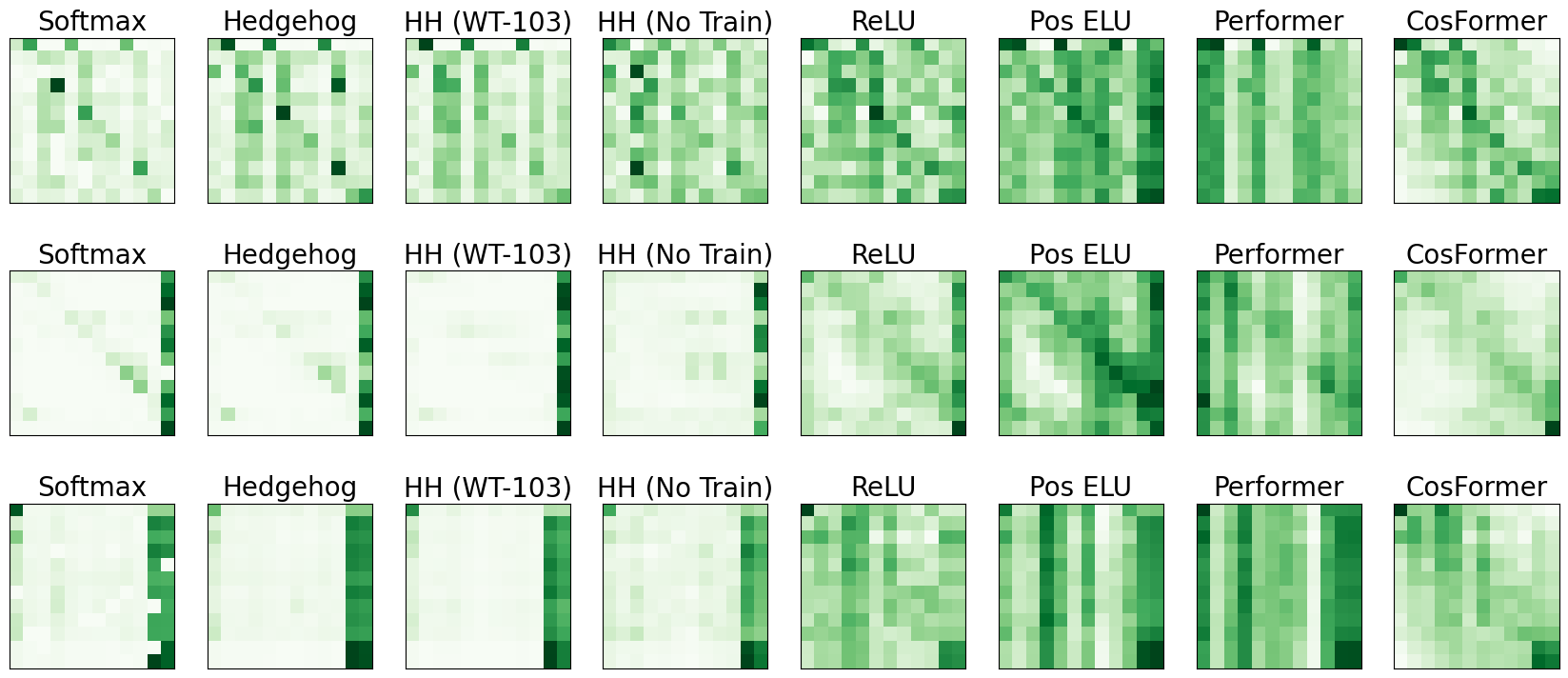}
    \caption{BERT attention visualizations for CoLA. Head 6; 1st, 6th, and 12th layers.}
    \label{fig:enter-label2}
    \vspace{-0.35cm}
\end{figure}

\begin{figure}
    \centering
    \includegraphics[width=0.95\linewidth]{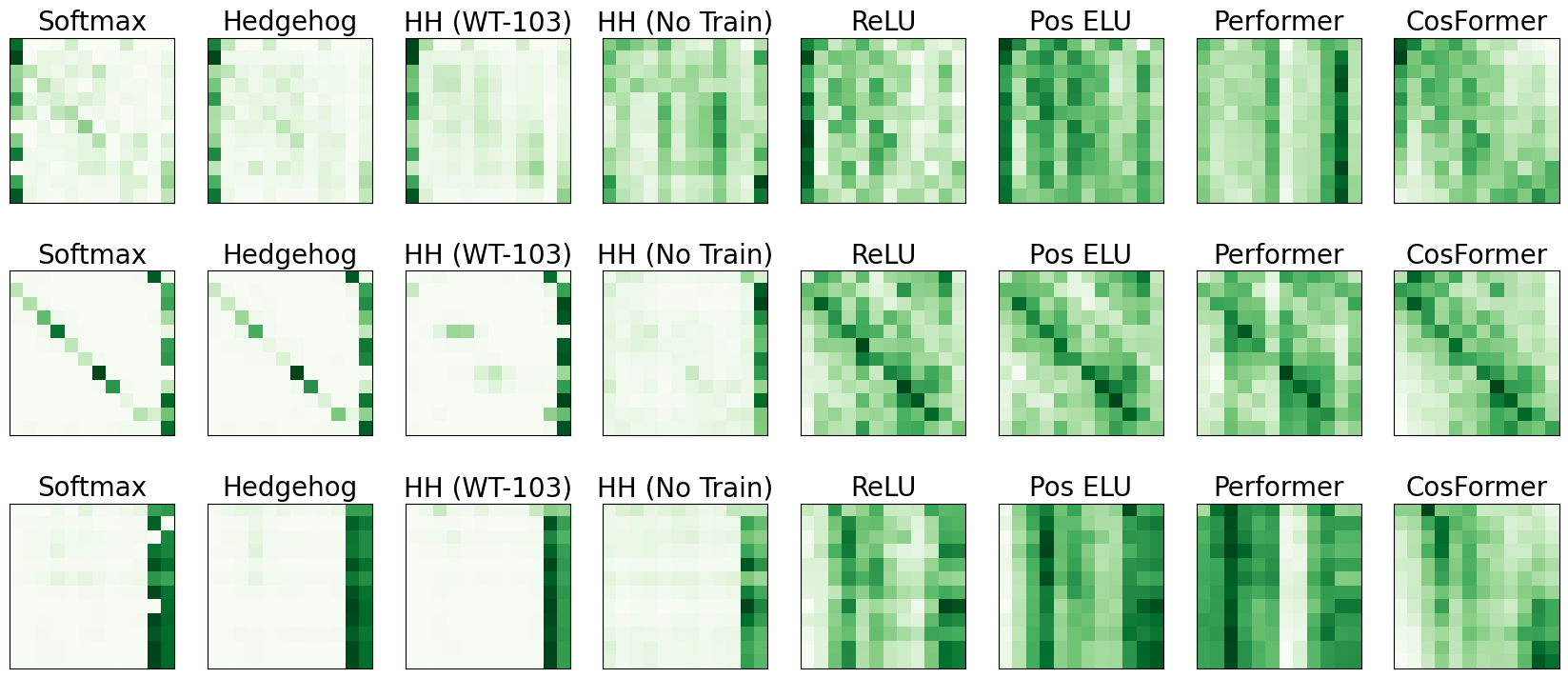}
    \caption{BERT attention visualizations for CoLA. Head 12; 1st, 6th, and 12th layers.}
    \label{fig:enter-label3}
    \vspace{-0.35cm}
\end{figure}

\newpage
\begin{figure}
    \centering
    \includegraphics[width=0.95\linewidth]{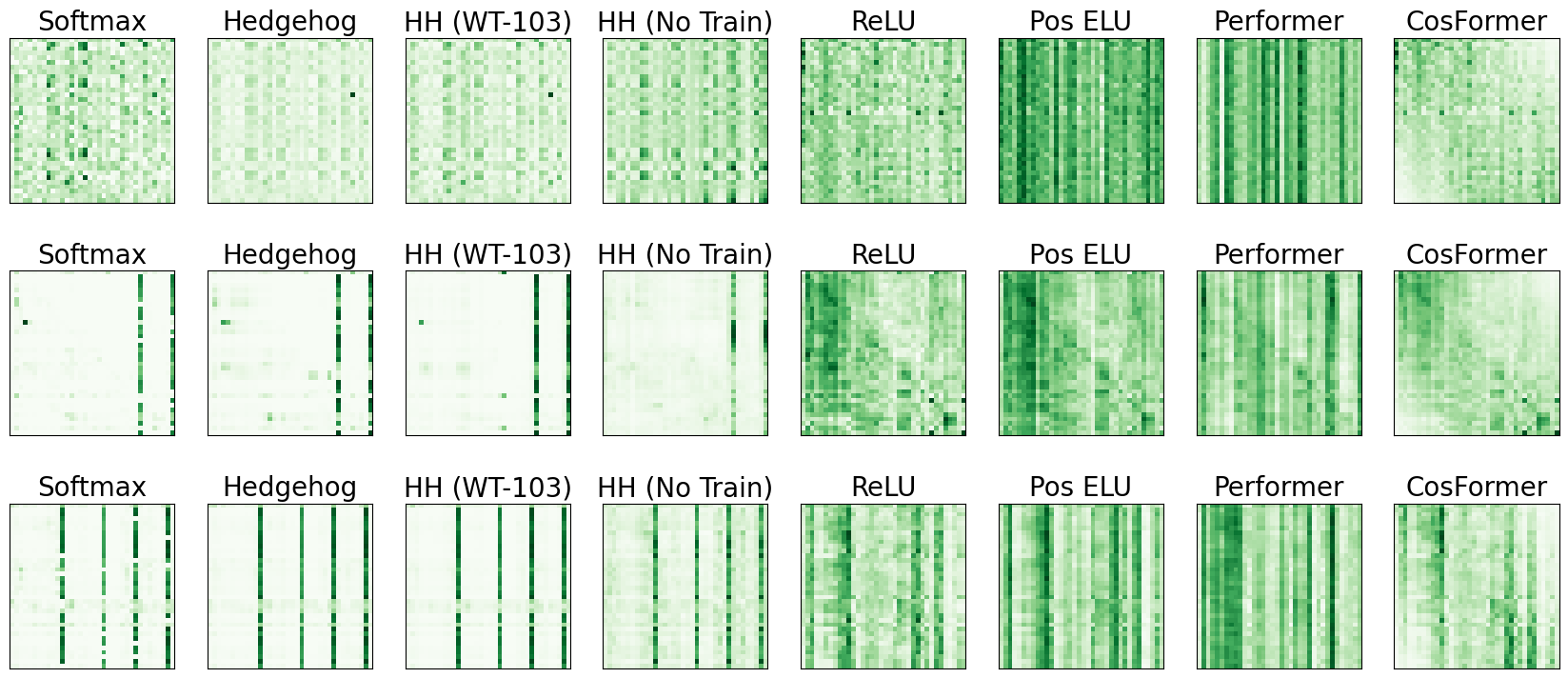}
    \caption{BERT attention visualizations for RTE. Head 0; 1st, 6th, and 12th layers.}
    \label{fig:enter-label4}
    \vspace{-0.35cm}
\end{figure}

\begin{figure}
    \centering
    \includegraphics[width=0.95\linewidth]{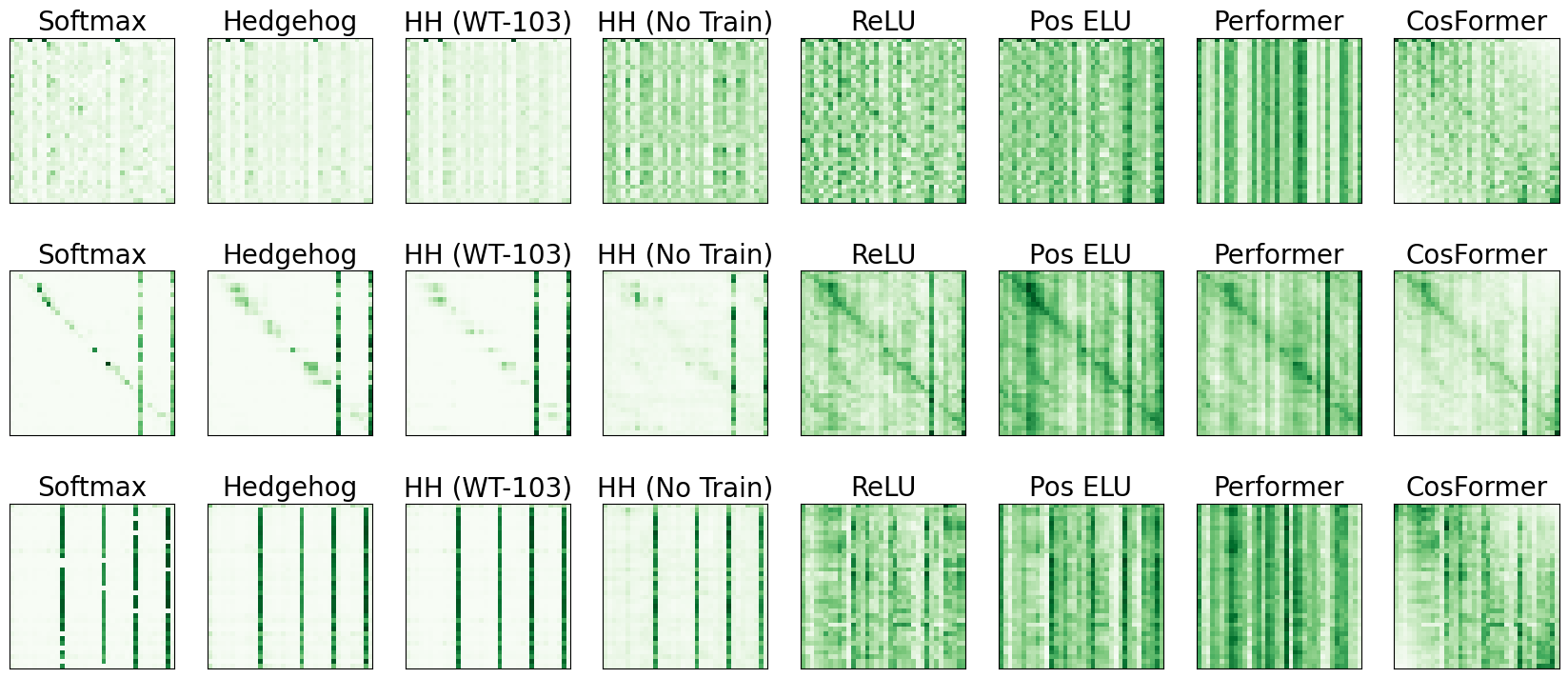}
    \caption{BERT attention visualizations for RTE. Head 6; 1st, 6th, and 12th layers.}
    \label{fig:enter-label5}
    \vspace{-0.35cm}
\end{figure}

\begin{figure}
    \centering
    \includegraphics[width=0.95\linewidth]{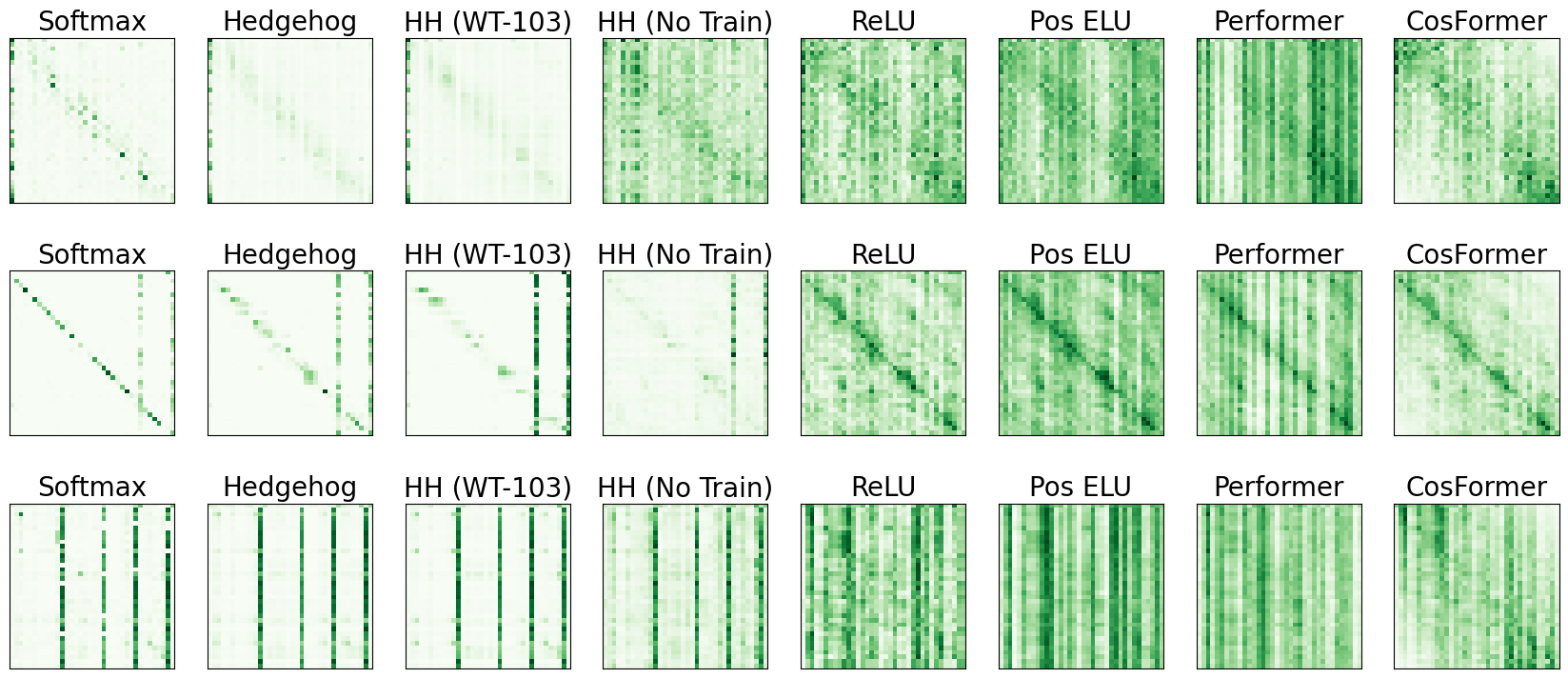}
    \caption{BERT attention visualizations for RTE. Head 12; 1st, 6th, and 12th layers.}
    \label{fig:enter-label6}
    \vspace{-0.35cm}
\end{figure}

\newpage
\begin{figure}
    \centering
    \includegraphics[width=0.95\linewidth]{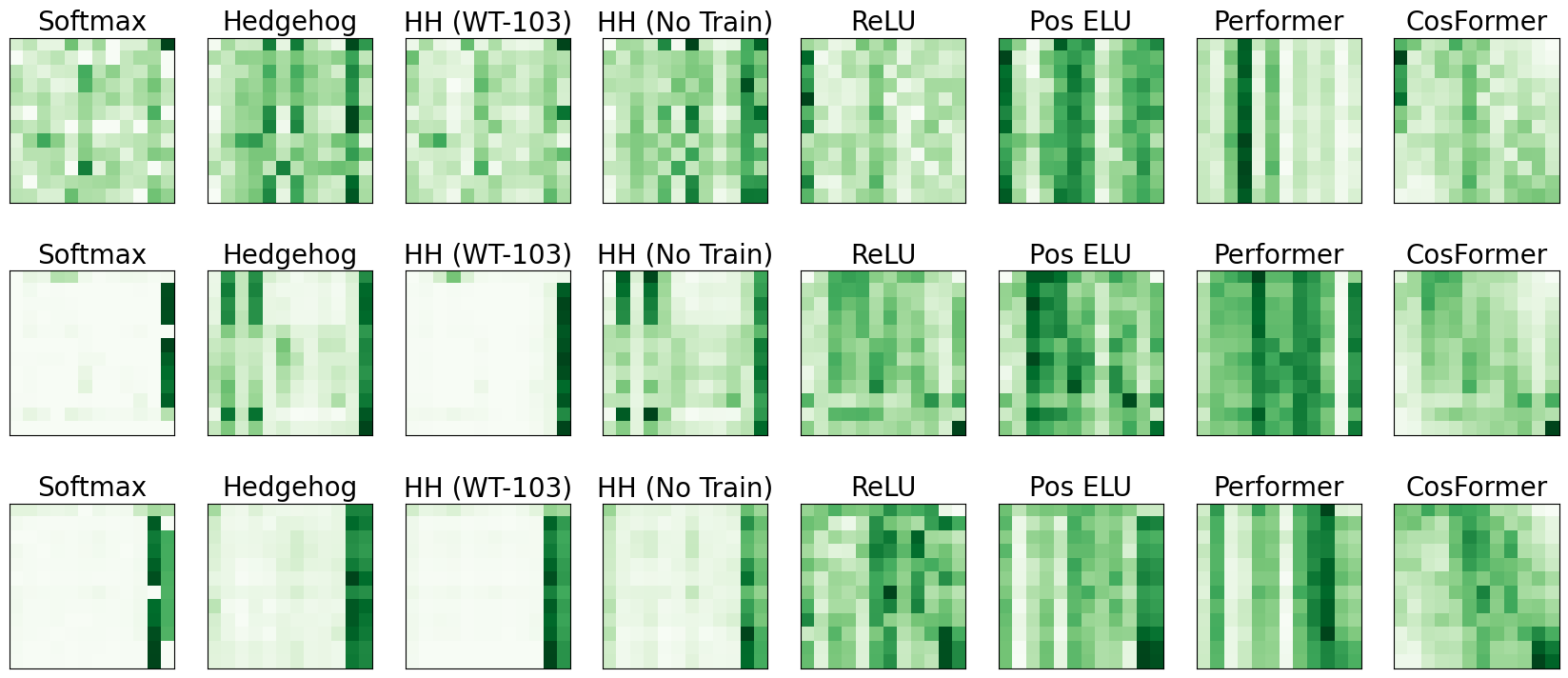}
    \caption{BERT attention visualizations for SST2. Head 1; 1st, 6th, and 12th layers.}
    \label{fig:enter-label7}
    \vspace{-0.35cm}
\end{figure}

\begin{figure}
    \centering
    \includegraphics[width=0.95\linewidth]{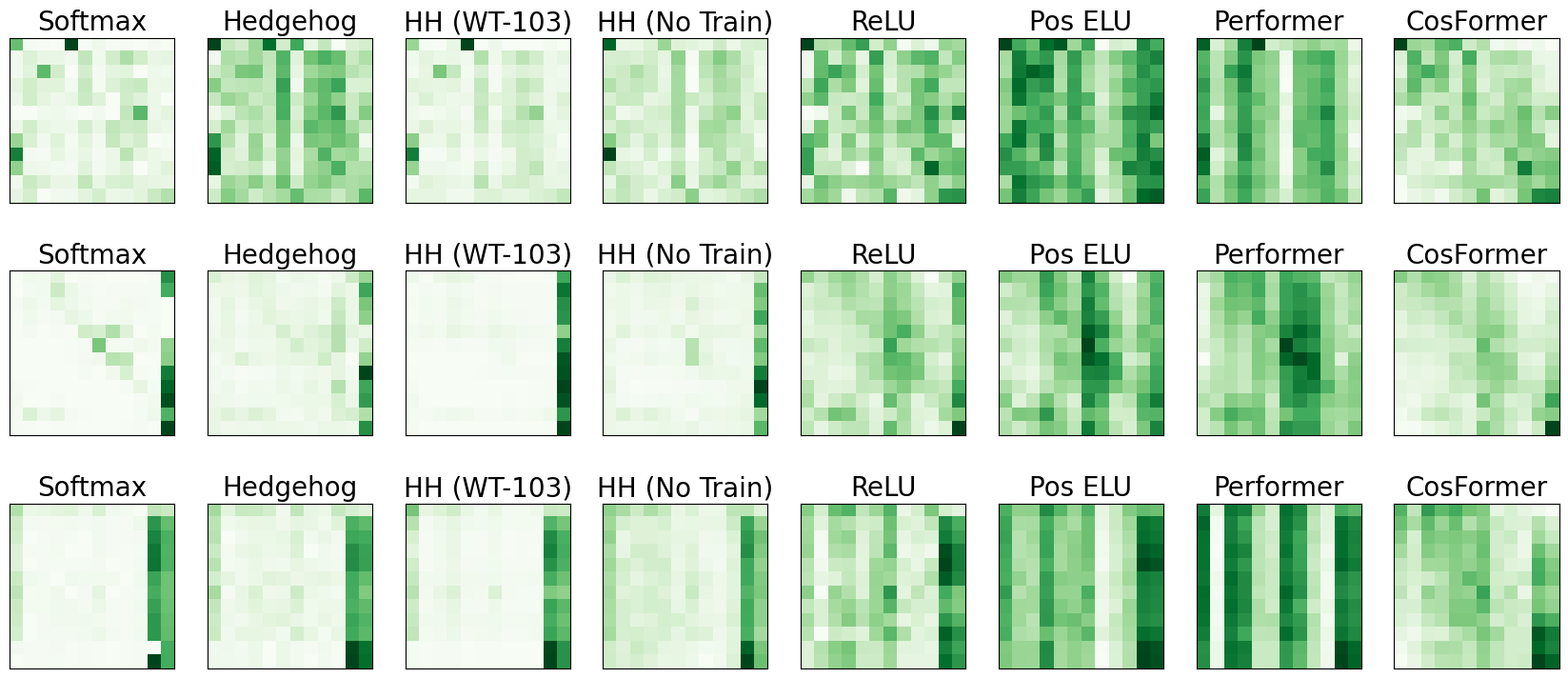}
    \caption{BERT attention visualizations for SST2. Head 6; 1st, 6th, and 12th layers.}
    \label{fig:enter-label8}
    \vspace{-0.35cm}
\end{figure}

\begin{figure}
    \centering
    \includegraphics[width=0.95\linewidth]{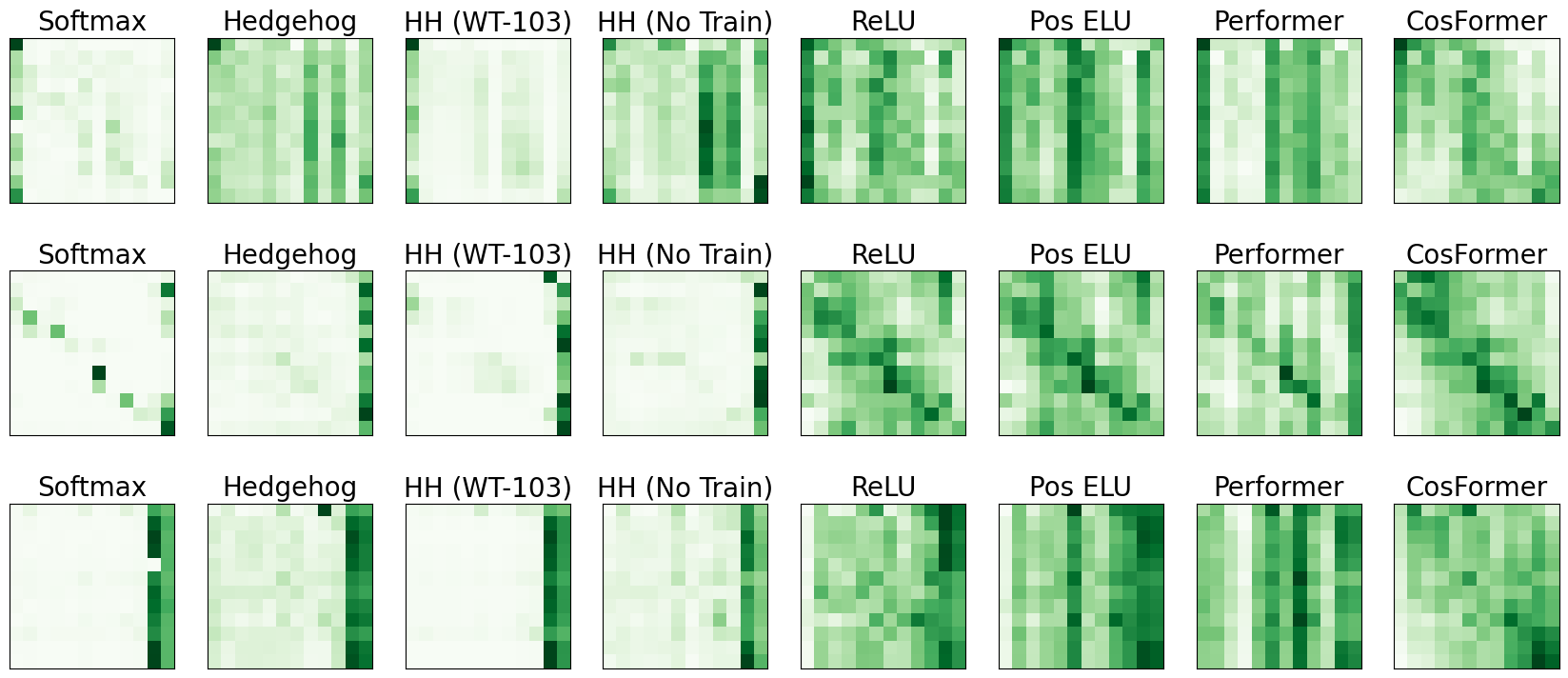}
    \caption{BERT attention visualizations for SST2. Head 12; 1st, 6th, and 12th layers.}
    \label{fig:enter-label9}
    \vspace{-0.35cm}
\end{figure}


\end{document}